\documentclass[lettersize,journal]{IEEEtran}
\IEEEoverridecommandlockouts
\usepackage{amsmath,graphicx}
\usepackage[utf8]{inputenc}

\usepackage[T1]{fontenc}
\usepackage{siunitx}
\usepackage{orcidlink}
\usepackage{nicefrac}
\usepackage{gensymb}
\usepackage{multirow}
\usepackage{pifont}
\usepackage[autostyle]{csquotes}

\usepackage{subcaption}

\usepackage{tcolorbox}
\usepackage{xcolor}
\usepackage[export]{adjustbox}
\graphicspath{ {./images/sim/} }
\usepackage{breqn}

\usepackage{color}
\definecolor{arrowblue}{rgb}{0,0,0.8}
\usepackage{acronym}
\usepackage{multirow}
\usepackage{array}
\usepackage{amssymb}
\usepackage{algorithm}
\usepackage{algpseudocode}
\usepackage[utf8]{inputenc}
\usepackage[T1]{fontenc}
\usepackage{url}
\usepackage{graphicx}
\usepackage{subcaption}
\usepackage{rotating}
\usepackage{tcolorbox}
\usepackage{url}

\usepackage{cite} 
\usepackage[export]{adjustbox}
\usepackage{graphicx,color}
\usepackage{hyperref}
\usepackage{multirow}
\acrodef{HDR}{high dynamic range}

\begin{document}
\title{Saturation-Aware Space-Variant Blind Image Deblurring}
\author{
   \hspace{-0.5cm}Muhammad Z. Alam\orcidlink{https://orcid.org/0000-0002-0114-8248}, Senior Member, IEEE,\thanks{Muhammad Z. Alam, Faculty of Computer science, University of New Brunswick, (e-mail: muhammad.alam@unb.ca)}
  \and
  Larry Stetsiuk, \thanks{Larry Stetsiuk, Department of Computer Science, Brandon University  (e-mail: stetsil37@brandonu.ca)}
\and
  Arooba Zeshan\orcidlink{https://orcid.org/0000-0001-7300-5804}, \thanks{Arooba Zeshan, Department of Applied Computer Science, University of Winnipeg  (e-mail: A.zeshan@uwinnipeg.ca)}
      
}

\maketitle

\begin{abstract}
This paper presents a novel saturation-aware space-variant blind image deblurring framework designed to address challenges posed by saturated pixels in deblurring under high dynamic range and low-light conditions. The proposed approach effectively segments the image based on blur intensity and proximity to saturation, leveraging a pre-estimated Light Spread Function (LSF) to mitigate stray light effects. By accurately estimating the true radiance of saturated regions using the dark channel prior, our method enhances the deblurring process without introducing artifacts like ringing. Experimental evaluations on both synthetic and real-world datasets demonstrate that the framework improves deblurring outcomes across various scenarios, showcasing superior performance compared to state-of-the-art saturation-aware and general-purpose methods. This adaptability highlights the framework's potential integration with existing and emerging blind image deblurring techniques.
\end{abstract}
\section{Introduction}
Blind image deblurring, a pivotal task in the realm of image processing, aims to restore sharpness to images that have been degraded by unknown blur, often caused by camera motion or focus issues. This endeavor becomes particularly challenging when dealing with images captured in low-light conditions or scenarios with high dynamic range, where pixel saturation often obscures crucial information necessary for effective deblurring .\\
It disrupts the linear relationship assumed in most deblurring algorithms between the blurred image and the unknown sharp image convolved with the blur kernel. Consequently, the saturated pixels do not adhere to the convolution model that underpins conventional deblurring methods, rendering standard kernel estimation techniques ineffective. Furthermore, saturated regions often introduce artifacts in the deblurring process, such as ringing effects, which further degrade the quality of the restored image. These technical challenges necessitate specialized approaches that can either recover or effectively circumvent the lost information in saturated areas to achieve successful image deblurring.\\
The optical system in a typical camera often induces wavefront alterations such as diffraction, which is particularly pronounced at the juncture of highly saturated (overexposed) and dark regions. Intense light emanating from saturated zones, upon interaction with these intricate lens elements, generates diffraction effects. \\
These optical interactions cause a spatial redistribution of photons, channeling them into adjacent, darker pixels. This results in a blooming artifact, characterized by a smearing of light into areas that ideally should remain dark. Scattering (reflecting or refracting) of the light from the surface of the lens is a major limitation of any camera system as it can substantially reduce the dynamic range of captured images or the loss of important details and features.\\
In an ideal scenario, a point light source that is in focus should only illuminate a single pixel. However, in reality, due to the scattering and reflections of light inside the camera lens and body, the light also reaches other pixels on the sensor.\\
Additionally, the blurring phenomenon leads to the spread of light from illuminated regions into adjacent darker areas, a process fundamentally governed by the convolution operation. Technically, when an image undergoes blurring, whether due to camera shake, defocus, or other factors the intensity values of each pixel become a weighted average of their own and neighboring pixels' intensities. \\
This averaging effect is determined by the convolution of the image with a Point Spread Function (PSF), which characterizes the nature of the blur. In regions where bright and dark areas are juxtaposed, this convolution process results in light 'leaking' from the bright (saturated or high-intensity) areas into adjacent darker regions.\\
The phenomena of blurring, and scattering,  particularly in the presence of larger saturated regions, contribute significantly to the alteration of pixel intensity values, imparting additional light to other pixels. Introduced initially in the context of image dehazing, the dark channel prior is predicated on the observation that in most natural, non-saturated images, there are some pixels with very low-intensity values, approaching zero, in at least one color channel and therefore, intensity values at these pixels, in a blurred version of the image with significant saturated pixels, can be considered as the approximated amount of imparted light at these pixels.\\
\begin{figure*}[h]
\centering
\includegraphics[trim={10 185 6 40},clip,width=6.5in,height=5.0in,keepaspectratio]{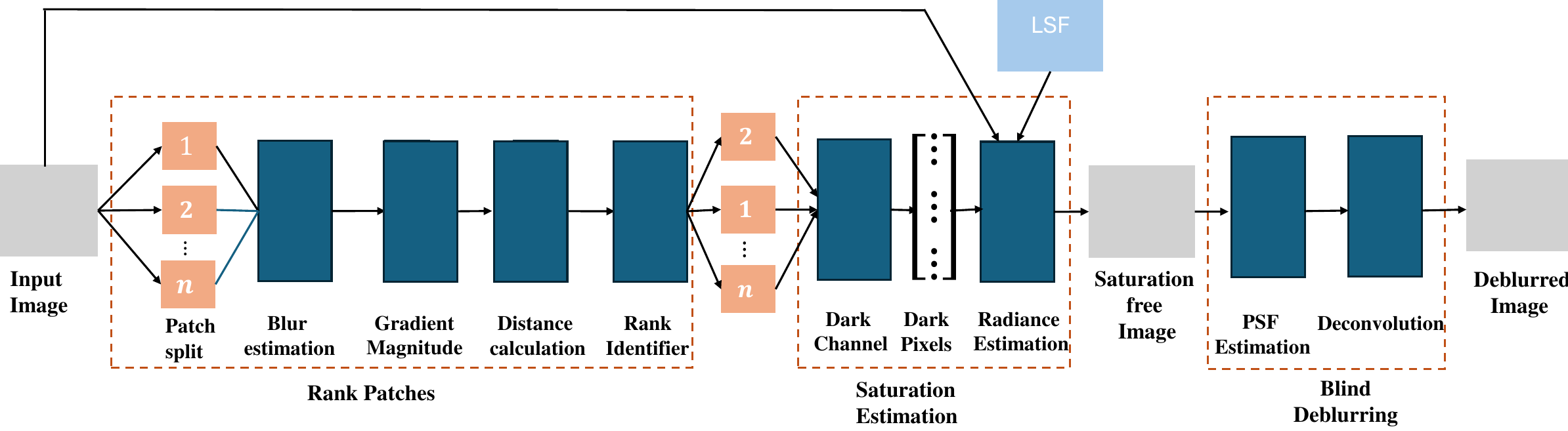}
\caption{Illustration of the proposed framework for saturation-aware space-variant blind image deblurring.}
\label{illustration}
\end{figure*}
In this paper, we present an innovative image deblurring framework, highlighted in Figure \ref{illustration}, that employs a strategic segmentation approach focusing on the level of blur and the proximity to saturated regions. This approach is specifically crafted to identify dark pixels that are minimally affected by blur, while also capitalizing on the light scattering from brighter regions. We ascertain that the light reaching these pixels predominantly emanates from bright sources, which are typically saturated in captured images. Hence, we can approximate the radiance of these saturated pixels by gauging the light gathered by the dark pixels, which is influenced by the luminance of bright, saturated areas.\\
Following the determination of the true intensity of these saturated pixels, the next phase in image restoration involves the blind estimation of PSF amidst stray light. For an effective PSF estimation, we employ a pre-calculated Light Spread Function (LSF) that accounts for the reflected component of light to the overall radiance of a scene. The final stage of restoring the blur-degraded image entails the deconvolution process using the estimated PSF.\\
The saturation estimation technique has been briefly presented in the context of glare reduction in \cite{9688911} and is limited to scenarios without the presence of blur. 
The major contributions of this paper are summarized as follows:
\begin{itemize}
    \item This paper introduces a blind image deblurring framework specifically designed to handle the presence of saturated pixels.
    \item  This paper addresses stray light mitigation in the presence of blur through a pre-estimated Light Spread Function (LSF).
    \item We test and evaluate the single-camera LSF on publicly available datasets captured from various camera setups.
\end{itemize}
Our code is available at \href{https://github.com/ZeshanAlam/Saturation-Aware-Space-Variant-Blind-Image-Deblurring}{\color{pink}Image_deblurring}.\\
\section{Related Work}
Blind image deblurring, a critical challenge in image restoration, seeks to recover clear images from blurred inputs by estimating an unknown blur kernel, a task made even more complex in scenarios involving saturated pixels. In recent years, image deblurring has been successfully addressed by several different methods \cite{10047966,ALAM2019,bai2018graph, J.sun, S.Nah, SunEvent, laroche2024fast, wang2023ddnm, chen2021blind, pan2016blind, chen2021learning, chen2020oid, cho2011handling, hu2014deblurring,kong2023efficient}. Zhang \emph{et al.} \cite{10047966} proposed Infwide, a Wiener deconvolution-based network designed for non-blind deblurring in low-light environments. This method enhances both image and feature space representations by employing the Wiener deconvolution filter, which minimizes noise and amplifies details, improving deblurring performance even under extreme lighting conditions. Alam \emph{et al.} \cite{ALAM2019} tackled space-variant blur kernel estimation by introducing a clustering approach to enhance the accuracy of deblurring.\\  In \cite{bai2018graph} a graph-based framework, leverages the graph representation of image patches to effectively model local and non-local dependencies by integrating structural priors and edge-preserving constraints. The method enhances robustness in handling complex blur kernels while preserving image details. In \cite{J.sun} and \cite{S.Nah}, the focus is on addressing non-uniform motion blur and dynamic scene deblurring using CNNs and multi-scale architectures, respectively. Sun \emph{et al.} \cite{SunEvent} utilized event-based fusion combined with cross-modal attention to mitigate blur, showcasing the effectiveness of combining different data modalities.\\
The fast diffusion EM model \cite{laroche2024fast} uses a diffusion process to solve blind inverse problems like deblurring. By modeling the noise and iteratively refining the image, this approach efficiently recovers sharp details without requiring prior knowledge of the blur kernel. Wang \emph{et al.} \cite{wang2023ddnm} developed a zero-shot image restoration method using a denoising diffusion null-space model.
Most of the blind and non-blind deconvolution methods designed for image deblurring do not explicitly account for saturation, but there are a few methods that do consider saturation in images or the night conditions in which lights typically get saturated \cite{chen2021blind, pan2016blind, chen2021learning, chen2020oid, cho2011handling, hu2014deblurring}. Saturation often occurs in high dynamic range environments, where pixels in the image lose information due to excessive brightness. The deblurring method in \cite{chen2021blind} integrates a saturation detection mechanism that identifies saturated regions and discards them during the blur kernel estimation. The algorithm then reconstructs the blurred image by estimating the missing information from the unsaturated areas.
Pan \emph{et al.} \cite{pan2016blind} utilized the dark channel prior for blind image deblurring, particularly effective in low-light situations. 
This prior is used to estimate the transmission map and the atmospheric light, helping the algorithm to distinguish between blurred and sharp regions. 
In their follow-up work Chen \emph{et al.} \cite{chen2021learning},  tackled non-blind deblurring of night-time images, a scenario characterized by extremely low-light environments, where both motion blur and noise significantly affect image quality. Their method employs a deep learning-based model that incorporates both a deblurring and denoising network. 
In \cite{chen2020oid}, and \cite{cho2011handling} outlier identifying and discarding methods for blind image deblurring target the removal of outliers that often distort the blur kernel estimation process. 
In \cite{hu2014deblurring} a streak-aware deblurring model that detects and models these streaks separately from the blur kernel is proposed. By isolating and handling these artifacts, the model reduces their impact on the overall deblurring process, resulting in clearer, artifact-free images.  Blind-deblurring scheme \cite{zhang2022pixel} screens intermediate images via a Bayes posterior to exclude saturated/outlier pixels that would bias kernel estimation, improving robustness on images with saturations and large blurs. Nighttime non-blind method \cite{shu2024deep} combines a deep prior with saturated-pixel handling (pixel-stretching mask, segment mask, and a saturation-awareness mechanism) to cope with varying saturation levels.\\
 Recent advancements in image deblurring have increasingly leveraged deep learning architectures that focus on frequency-domain analysis, hierarchical modeling, and generative priors. Mao et al. \cite{LoFormer} proposed LoFormer, a transformer-based architecture that emphasizes local frequency information to better capture spatially variant blur characteristics. Their approach achieves high performance by enhancing the model's ability to preserve local structure and frequency consistency during restoration. Similarly, Li et al. \cite{li2024hierarchical} introduced a hierarchical wavelet-guided diffusion model, which exploits multi-scale frequency information through wavelet transforms, effectively guiding the denoising process in a coarse-to-fine manner.\\
Diffusion-based methods have also gained prominence for their ability to model complex data distributions. Chen et al. \cite{chen2024hierarchical} presented a Hierarchical Integration Diffusion Model (HIDM) that integrates different levels of feature representations within a diffusion process to improve realism in deblurred outputs. Meanwhile, unsupervised approaches have been explored to overcome limitations in ground-truth availability. Chen et al. \cite{chen2024unsupervised} developed a self-enhancement mechanism for unsupervised blind image deblurring, enabling the model to learn effective priors directly from corrupted data. Mao et al. \cite{mao2024adarevd} proposed AdaReVD, an adaptive reversible decoder architecture that introduces early-exiting mechanisms at the patch level, allowing flexible and efficient inference while maintaining competitive performance.
These recent methods represent state-of-the-art advancements in both supervised and unsupervised deblurring. \\
Most existing deblurring methods, including traditional blind deconvolution, CNN-based architectures, and recent transformer-based models, are not designed to handle saturated pixels. A few methods \cite{chen2021blind, pan2016blind, chen2021learning, chen2020oid, cho2011handling, hu2014deblurring} consider saturation, but typically do so by excluding saturated regions during blur kernel estimation or treating them as outliers. In the proposed technique, we adopt a different approach by estimating the true radiance of saturated pixels rather than avoiding saturation altogether.  Beyond saturation, deblurring must also contend with measurement noise; dual-noise robustness \cite{zhang2025reliable} suggests modeling concurrent noise sources/uncertainty, analogous to treating saturated pixels as corrupted measurements. Parameter choices for saturation (e.g., thresholds/priors) can be cast as conditional adaptation; conditional prompt-tuning \cite{zhang2025} motivates input-conditioned parameterization that adjusts these settings from image content. Finally, blur and saturation vary across scales; \cite{10308720} highlights multi-scale checks for distribution shift, informing region-wise handling in space-variant deblurring. \color{black} To predict the intensity of saturated regions, we utilize the dark channel prior. However, the presence of blur complicates the saturation estimation process. The proposed method leverages the space-variant nature of the blur to improve the accuracy of saturation estimation, enabling more precise radiance prediction for saturated areas.\color{black}

\section{Saturation aware space-variant deblurring framework (SASVD)}
The proposed saturation-aware space-variant blind image deblurring framework consists of two key stages, presented in Algorithm \ref{algo1}, aimed at addressing the challenges in image deblurring in the presence of saturation. The first stage focuses on accurately estimating the saturated regions by leveraging the dark channel prior, enabled by the separation of light contributions from scattering and blurring effects. Once the true radiance of these regions is recovered, the second stage applies a blind image deblurring algorithm to restore the sharpness of the entire image while accounting for the estimates of the saturated regions. This two-step approach ensures precise handling of saturation and enhances the effectiveness of deblurring specifically in spatially variant blur scenarios.
 \begin{algorithm}
\caption{Saturation-aware space-variant blind deblurring}
\begin{algorithmic}[1]
\State Initialize: 
\State $B \gets \text{Input Blurred Image}$
\State $I \gets \text{LSF}$
\State $P_i = \text{DivideImage}(B)$
\State $S_i, B_i = \text{EstimateBlurAndSharpness}(P_i)$
\State $R_i = \text{CalculateProximityToSaturation}(P_i)$
\State $\text{RankedPatches} = \text{RankPatches}(P_i, S_i, B_i, R_i)$
\State $D = \text{SelectDarkPixels}(\text{RankedPatches})$
\State $S_{\text{estimate}} = \text{EstimateSaturation}(D)$
\State $B_{\text{modified}} = \text{ReplaceSaturatedPixels}(B, S_{\text{estimate}})$
\State $X = \text{BlindDeblurring}(B_{\text{modified}}, K)$
\State Output: $X \gets \text{Deblurred Image}$
\end{algorithmic}
\label{algo1}
\end{algorithm}
\subsection{Deblurring in the presence of saturation}
Deblurring is a process aimed at restoring a sharp image from a blurred one by estimating the blur kernel and reversing its effects. This process typically relies on the assumption of a linear relationship between the observed blurred image \( Y \) and the underlying sharp image \( X \), represented mathematically as:
\begin{equation}
Y = B \ast X + n
\end{equation}
where \( B \) is the blur kernel, and \( n \) is the additive noise. However, this linear assumption breaks down in the presence of saturated regions, where pixel intensity reaches its maximum limit due to extreme brightness (e.g., vehicle headlights or sunlight). These saturated regions violate the linear model because their pixel values are clipped, leading to significant information loss and incorrect blur estimations during deblurring.\\
In an ideal imaging scenario, a bright light source, such as a car headlight or the sun, should illuminate only the pixels directly corresponding to the light's position in the image. However, due to internal reflections and scattering within the camera system, this light spreads to neighboring pixels, causing an unintended increase in brightness in areas surrounding the light source. This spread of light is influenced by what can be described as the \textit{Light Spread Function (LSF)}. The LSF represents how light from a point source is dispersed across the image sensor due to these internal camera effects. This can be mathematically represented as:
\begin{equation}
Y' = I \ast X + n
\label{intesnity}
\end{equation}
where, \( Y'\) is the observed image, \( I \) models the light scattering and internal reflections, and
\( n \) is the noise. The result of this convolution is that light, instead of being confined to its original location, spreads into adjacent pixels, causing areas around bright light sources to become brighter than they should be.
\subsection{The Combined effect}
\begin{figure*}[!h]
\centering
\begin{subfigure}{1.70in}
{\includegraphics[width=1.70in,height=1.70in,keepaspectratio]{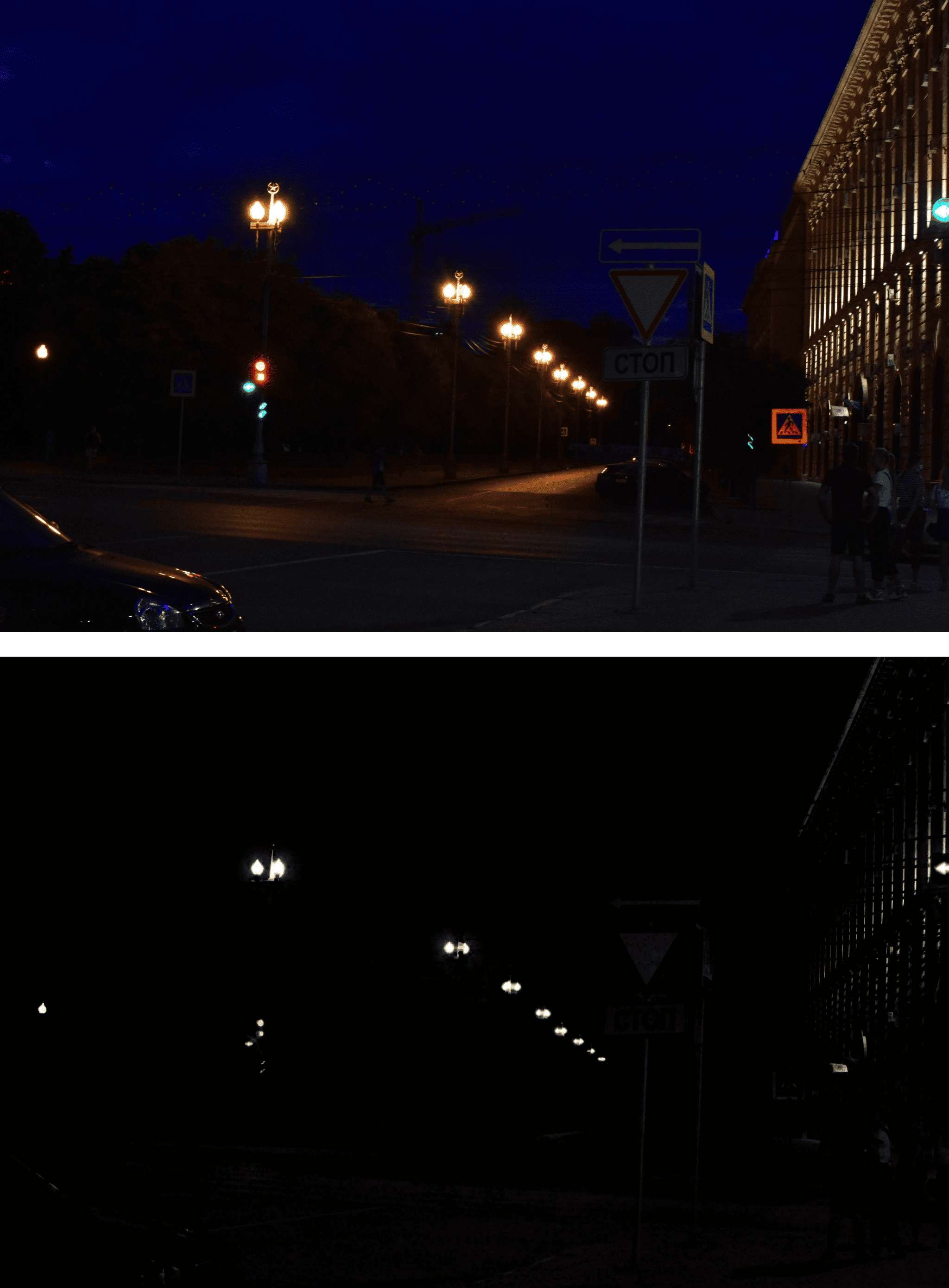}}
  \caption{42_NIKON-D3400-35MM_S}

\end{subfigure}
\begin{subfigure}{1.70in}
{\includegraphics[width=1.70in,height=1.70in,keepaspectratio]{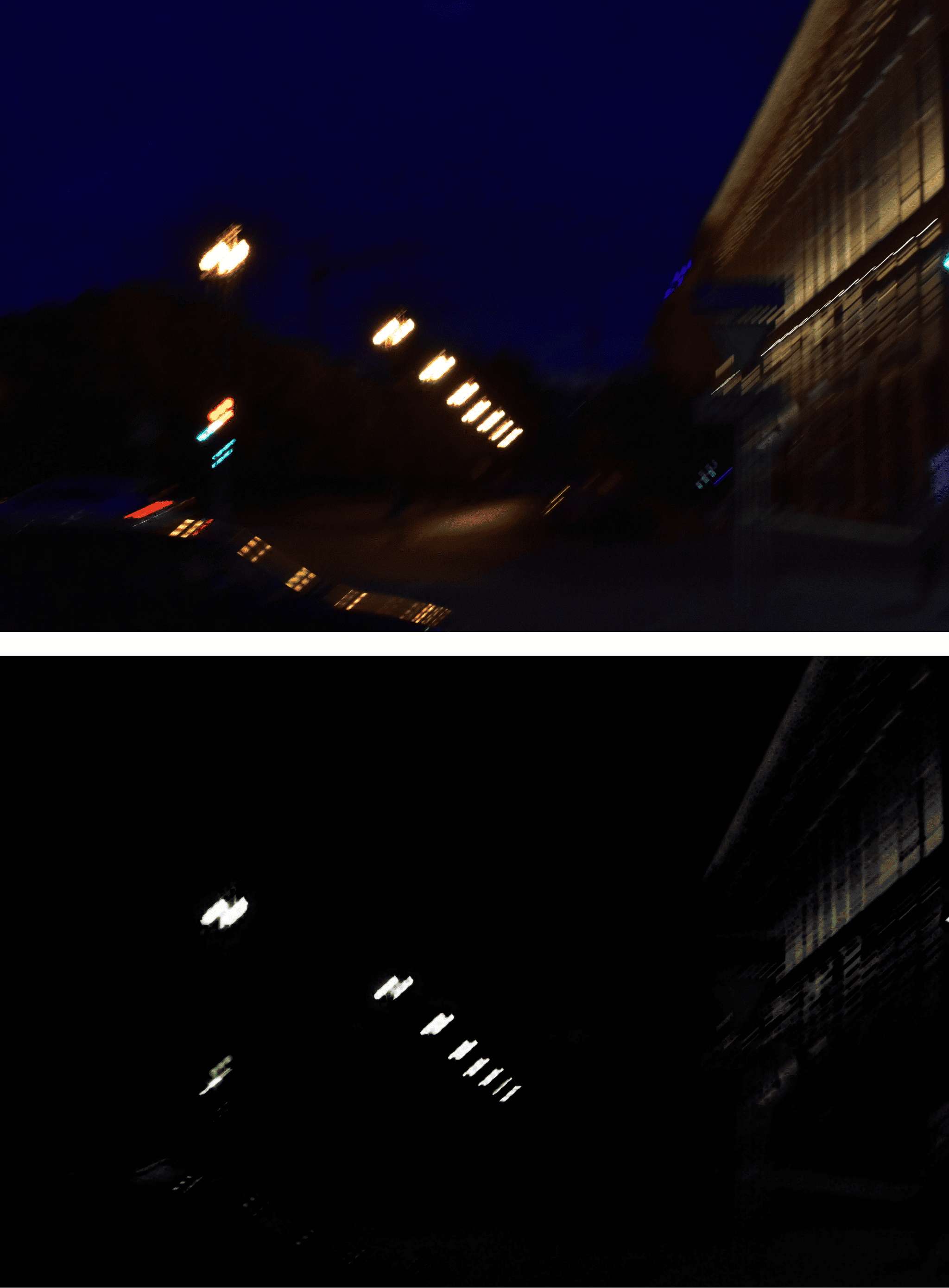}}
\caption{42_NIKON-D3400-35MM_M}

\end{subfigure}
\begin{subfigure}{1.70in}
{\includegraphics[width=1.70in,height=1.70in,keepaspectratio]{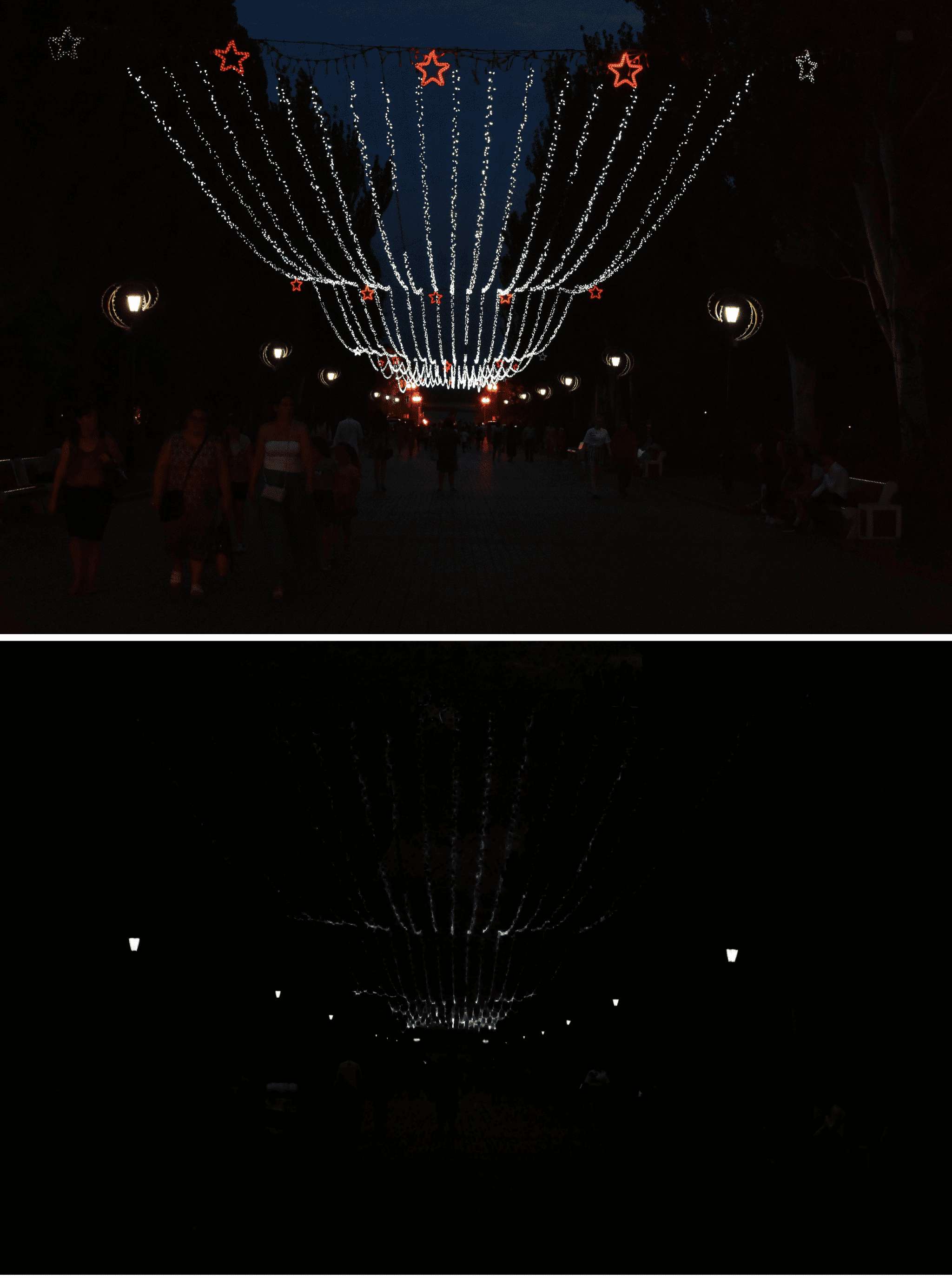}}
 \caption{53_NIKON-D3400-35MM_S}

\end{subfigure}
\begin{subfigure}{1.70in}
{\includegraphics[width=1.70in,height=1.70in,keepaspectratio]{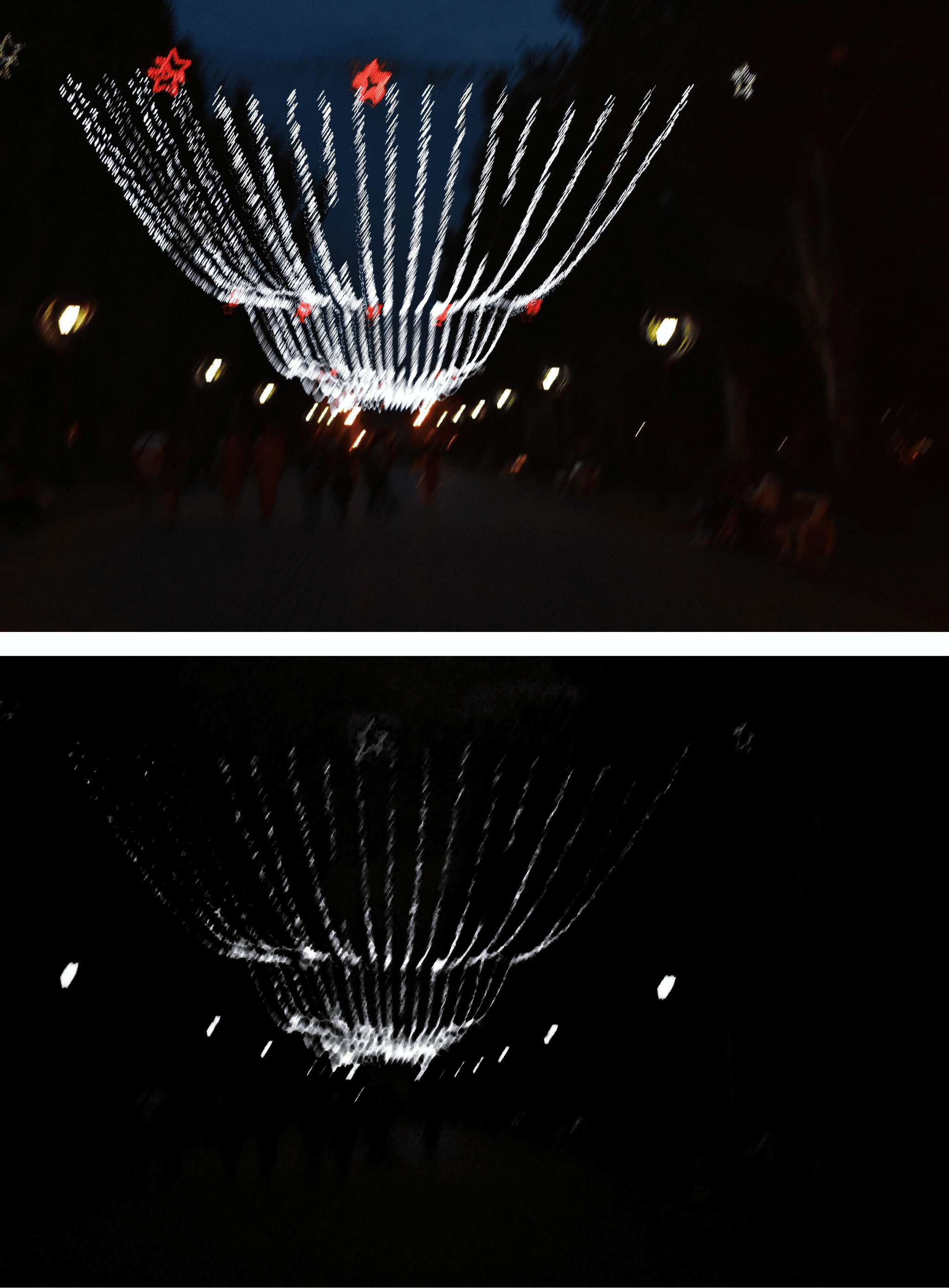}}
 \caption{53_NIKON-D3400-35MM_M}

\end{subfigure}
 \caption{Dark channel sparsity comparison on selected natural low-light (night-time) images from \cite{blur_dataset}. For each subfigure, the top image is the input regular image, and the bottom image is the corresponding dark channel (computed using a patch size of 5×5).
Two natural image pairs showing clear (left) and blurred (right) versions from the dataset, labeled as (a–d), illustrate dark channel degradation under real-world blur conditions.color{black}}
\label{blur-clear}
 \end{figure*}

Blurring affects the intensity distribution of an image by averaging pixel values in a local neighborhood, which increases the intensity of dark pixels.  Blur further complicates the already altered intensity distribution due to saturation in an image by averaging the intensities of the neighboring pixels. The presence of blur modifies the equation \ref{intesnity} as follows:
\begin{equation}
Y'' = (B \ast (X \ast I) + n)
\label{combined}
\end{equation}
In the presence of blur, \( X \)  is convolved with the LSF and then further convolved with the blur kernel \( B \), making it difficult to distinguish the true contributions of light to the neighboring pixels from each phenomenon, complicating the estimation of the true radiance of the saturated regions in  \( X \). We further support our analysis through experiments. Figure \ref{blur-clear} clearly demonstrates how increasing levels of blur complicate the estimation of saturation due to changes in the intensity of neighboring pixels. As blur spreads across an image, pixel values undergo an averaging effect, causing high-intensity saturated regions to diffuse into adjacent areas. This diffusion alters the local intensity distribution and makes it challenging to accurately distinguish true radiance values within saturated regions. Consequently, the process of estimating saturation becomes more complex, as the influence of blur must be decoupled from the scattered light contribution to obtain precise radiance estimates.
\subsection{Estimating Saturated Regions}
The observed image \( Y'' \) in equation \ref{combined}, consists of both saturated regions \( S \) and unsaturated regions \( U \), and can be expressed as a combination of these components:
\begin{equation}
Y'' = \left(B \ast \left(X_s \ast I\right)\right) + \left(B \ast \left(X_u \ast I\right)\right) + n
\end{equation}
To accurately estimate the radiance of saturated regions in such scenarios, we employ the dark channel prior. The dark channel prior is based on the observation that in most natural images, some pixels in non-saturated areas have very
low intensity in at least one color channel. These dark pixels are influenced by the spread of light from bright sources and can thus serve as indicators of how much light has been dispersed into neighboring regions.\\
For dark pixels \( D \) within the unsaturated regions \( U \), the intensity \( Y''(D) \) is influenced by the light spread from the saturated regions \( S \), modeled by the LSF, and can be expressed as:
\begin{equation}
Y''(D) = \left(B \ast \left(X_s \ast I\right)\right)(D) + \left(B \ast \left(X_u \ast I\right)\right)(D) + n
\label{dark}
\end{equation}
where, \( X_s(D) \) represents the true radiance of the saturated regions, and  \( X_u(D) \) represents the true intensity of the unsaturated regions.
The goal is to accurately estimate the true radiance of the saturated regions \( X_s \) by minimizing the difference between the observed intensity of dark pixels \( Y''(D) \) and the modeled values based on the LSF. However, since the PSF and \( X_s \) are unknown, this complicates the estimation process, making it challenging to achieve accurate optimization.
\subsection{ Identifying Minimally Blurred Patches}
In natural images, blur is often spatially variant due to factors such as depth of field, motion blur, and optical aberrations, resulting in regions of differing blur levels across the image. To effectively isolate the influence of scattered light described by the light spread function, it is essential to identify image patches with minimal blur and are situated close to regions of high saturation. This approach aims to maximize the influence of scattered light on dark pixels while minimizing contributions from blur. The selection process involves multiple stages  highlighted in Algorithm \ref{indetifypatch}, \color{black} including gradient-based sharpness evaluation, blur estimation via blind deconvolution, and a ranking scheme based on blur and spatial proximity.\\
\begin{algorithm}[ht]
\caption{Identify Patches with Low Blur and Distance from Saturated Region}
{Image $I$, patch size $p \times p$, saturation threshold $T_s$, blur threshold $\tau_b$, sharpness threshold $\tau_g$}
{Top $N$ reliable patches $\{P_i\}_{i=1}^N$}

\textbf{1.} Identify saturated pixels:\quad $\mathcal{S} = \{(x, y) \mid \max_c I(x, y, c) \geq T_s \}$ \\
\textbf{2.} Compute centroid of saturated region: 
\[
{c}_s = \frac{1}{|S_k|} \sum_{(x,y) \in S_k} (x, y), \quad \text{where } k = \arg\max_i |S_i|
\]
\textbf{3.} Divide image $I$ into overlapping patches $P_i$ of size $p \times p$ \\
\textbf{4.} For each patch $P_i$ with center $\mathbf{c}_i$:
\begin{itemize}
    \item[] a) Estimate blur kernel:\quad $K_i \leftarrow \texttt{deconvblind}(P_i, \text{initialPSF}, \text{iterations})$
    \item[] b) Compute blur score:\quad $b_i = \mathrm{Var}(K_i)$
    \item[] c) Compute sharpness:\quad $g_i = \sum \left( \nabla_x P_i^2 + \nabla_y P_i^2 \right)^{1/2}$
    \item[] d) Compute proximity to saturation:\quad $d_i = \|\mathbf{c}_i - \mathbf{c}_s\|_2$
\end{itemize}
\textbf{5.} Filter patches satisfying:\quad $b_i \leq \tau_b$ and $g_i \geq \tau_g$ \\
\textbf{6.} Rank filtered patches by ascending $b_i$ and ascending $d_i$
\textbf{7.} Select top $N$ patches from the ranked list
\label{indetifypatch}
\end{algorithm}
Given the observed image $Y'' \in \mathbb{R}^{m \times n \times c}$ (where $m, n$ denote the image dimensions and $c$ represents the color channels), the image is divided into patches of size $p \times p$. Let each patch be represented as $P_{r,c} \subset Y''$, where $r$ and $c$ denote the starting coordinates of the patch.\\
\textit{Sharpness Estimate:} To assess the sharpness of each patch, we compute the gradient magnitude of its grayscale version. Given the intensity values of a patch $P_{r,c}$, the gradients along the horizontal and vertical directions, $g_x$ and $g_y$, are computed as:
\[
g_x, g_y = \nabla P_{r,c}
\]
The local intensity gradient magnitude for each pixel is given by:
\[
G_{r,c} = \sqrt{g_x^2 + g_y^2}
\]
The total gradient magnitude for the patch, serving as a sharpness measure, is computed as:
\[
M_{r,c} = \sum_{i=1}^{p} \sum_{j=1}^{p} G_{r,c}(i, j)
\]
A patch is selected for further processing if $M_{r,c} > \text{sharpness threshold}$.\\
\textit{Blur Estimation via Blind Deconvolution:}
 For each patch $P_{r,c}$ in the image, a quantitative measure of local blur is obtained using blind deconvolution using MATLAB’s built-in deconvblind function. The estimated blur kernel $K_{r,c}$ is computed through an iterative deconvolution process using a predefined initial point spread function (PSF). The resulting kernel's effectiveness in capturing the blur is quantified by calculating the variance of the kernel values, denoted as  \text{var}($K_{r,c}$). Patches with lower variance values are considered more blurred, while patches with higher variance values indicate minimal blur.\\
To ensure the selection of minimally blurred patches with significant exposure to scattered light, only patches satisfying a low blur threshold and a high sharpness threshold are retained for further processing. The filtered patches are then ranked, first by their blur measure and subsequently by their proximity to the centroid of the saturated region, ensuring that the selected patches are both minimally blurred and spatially relevant to the saturation context.\\
To measure the proximity of patches to the saturated region, the centroid of the saturated region is first determined based on pixels exceeding a specified saturation threshold, with coordinates denoted as \( (x_c, y_c) \).  This is defined as the mean coordinate of all pixels whose intensity exceeds the saturation threshold $T_s$, i.e.,
\begin{align}
    \mathbf{c}_s &= \frac{1}{|\mathcal{S}|} \sum_{(x, y) \in \mathcal{S}} 
    \begin{bmatrix}
        x \\ y
    \end{bmatrix}, \\
    \mathcal{S} &= \left\{ (x, y) \ \middle| \ \max_c I(x, y, c) \geq T_s \right\}
\end{align}
For images with multiple dis-joint saturated regions, the centeroid is calculated for the dominant source i.e, with the highest number of saturated pixels in the connected saturated region.

\color{black}For each patch in the image, the center coordinates are computed as:
\begin{equation}
 Patch Center =\left( r + \frac{p}{2}, c + \frac{p}{2} \right),
\end{equation}
where \( (r, c) \) denotes the top-left coordinates of the patch and \( p \) is the patch size.
The Euclidean distance between the patch center and the saturated region's centroid is calculated using the formula:
\begin{equation}
d_{r,c} = \sqrt{\left(r + \frac{p}{2} - x_c\right)^2 + \left(c + \frac{p}{2} - y_c\right)^2}
\end{equation}
This distance serves as a measure of proximity, with smaller distances indicating closer patches to the saturated region's centroid. All patches satisfying the blur and gradient thresholds, set empirically  (e.g., $\tau_b = 0.003$) and keep changing between different experiments and datasets, are ranked based on the blur measure and proximity to the saturated region centroid. The top $N=8$ patches are then selected for further analysis.\\
\\
\color{black}
The intensity equation \ref{dark} for the minimally blurred patches simplifies to:
\begin{equation}
Y''_{P_{r,c}}(D) = \left(X_s \ast I\right)(D) + \left(X_{uP_{r,c}} \ast I\right)(D) + n
\label{blurfree}
\end{equation}
Here, since the blur \( B \) is negligible for patches \(P_{r,c}\) it is effectively removed from the equation \ref{dark}, and the estimation of \( X_s \) relies purely on the dark pixels from non-blurred patches \(P_{r,c}\) affected only by the LSF.\\
The resulting equation \ref{blurfree} can be optimized to accurately estimate the saturated regions X_s as follows:
\begin{equation}
\begin{split}
\textrm{argmin}_{X_s,({I\ast X_{up}})(D)} \left \|Y''_{up}(D) - (I \ast X_s)(D)-   {(I \ast X_{up})(D)} \right \|_2 \\
+  \lambda \left \| {I \ast X_{up}}(D) \right \|_1.     
\end{split}
\label{const1}
\end{equation}
This objective function estimates the true radiance of the saturated regions by leveraging the dark channel prior.  We jointly optimize the pixel values in both the saturated and unsaturated regions, denoted by $\mathbf{X}_s$ and $\mathbf{X}_{up}$. The optimization is carried out using an iterative update strategy, where the gradients of the objective are computed with respect to both $\mathbf{X}_s$ and $\mathbf{X}_u$, and updates are applied concurrently.
The objective is evaluated only on $D$ to emphasize stray-light evidence (rather than blur aggregation); in practice, we initialize from the minimally blurred patches and jointly optimize $X_s$ and $X_u$ (alternating/gradient updates with non-negativity and the $\lambda\!\left\|\,(I * X_{up})(D)\,\right\|_{1}$ term) until convergence.\color{black}
 This joint estimation ensures that the reconstructed values across saturated and unsaturated regions remain consistent with the global blur model and observed image constraints.
\color{black} To ensure that the estimated values are physically realistic, the following constraints are applied during the optimization process:
\begin{equation}
Y''_u - I \ast(X_s  + \tilde{X}_u)(u) \geq 0
\end{equation}
\begin{equation}
X_u(D) \geq 0
\end{equation}
\begin{equation}
((X_s \ast I) + Y''_u ) \oslash I\geq 0
\end{equation}
where $\tilde{X}_u$ is obtained by the deconvolution of $Y''_u$ with $I$. These constraints guarantee that the pixel values remain positive and that the estimated radiance values are consistent with the observed data. This approach ensures that the contributions from both the saturated and unsaturated regions are correctly accounted for, leading to more accurate results.\\
It is important to emphasize that the successful recovery of the saturation free image $Y'$ relies on two primary factors: the availability of ample dark pixels in regions of the input image that are non-degraded by spatially varying blur, which enables accurate estimation of the stray light despite the presence of noise, and the accuracy of the adopted  model in measuring the LSF. In scenarios of extreme saturation, where over half of the image is affected and dark pixels are not available in sufficient numbers, the effectiveness of the proposed algorithm may degrade. In such cases, the saturation estimation module becomes unreliable, and the framework essentially reverts to a non-saturation-aware deblurring process. However, we note that the utility of recovering such heavily saturated images is inherently limited, as significant portions of meaningful visual content are already irreversibly lost due to saturation.\color{black}
\subsection{Image deblurring}
Once the saturation-free image \( Y' \) is estimated, the second stage of the proposed framework involves utilizing any typical blind image deblurring algorithm to recover a blur-free image. In this paper, we primarily employ the method by Xu \emph{et al.}\cite{xu2010two}. This method was chosen due to its deconvolution-based approach, which offers a balance between computational efficiency and accuracy. Additionally, it is well established in the community and is publicly available, making it accessible and reliable for our deblurring needs.
\section{Light Spread Function Pre-Estimation}
\begin{figure}[h]
\centering
\includegraphics[width=2.0in,height=2.0in,keepaspectratio]{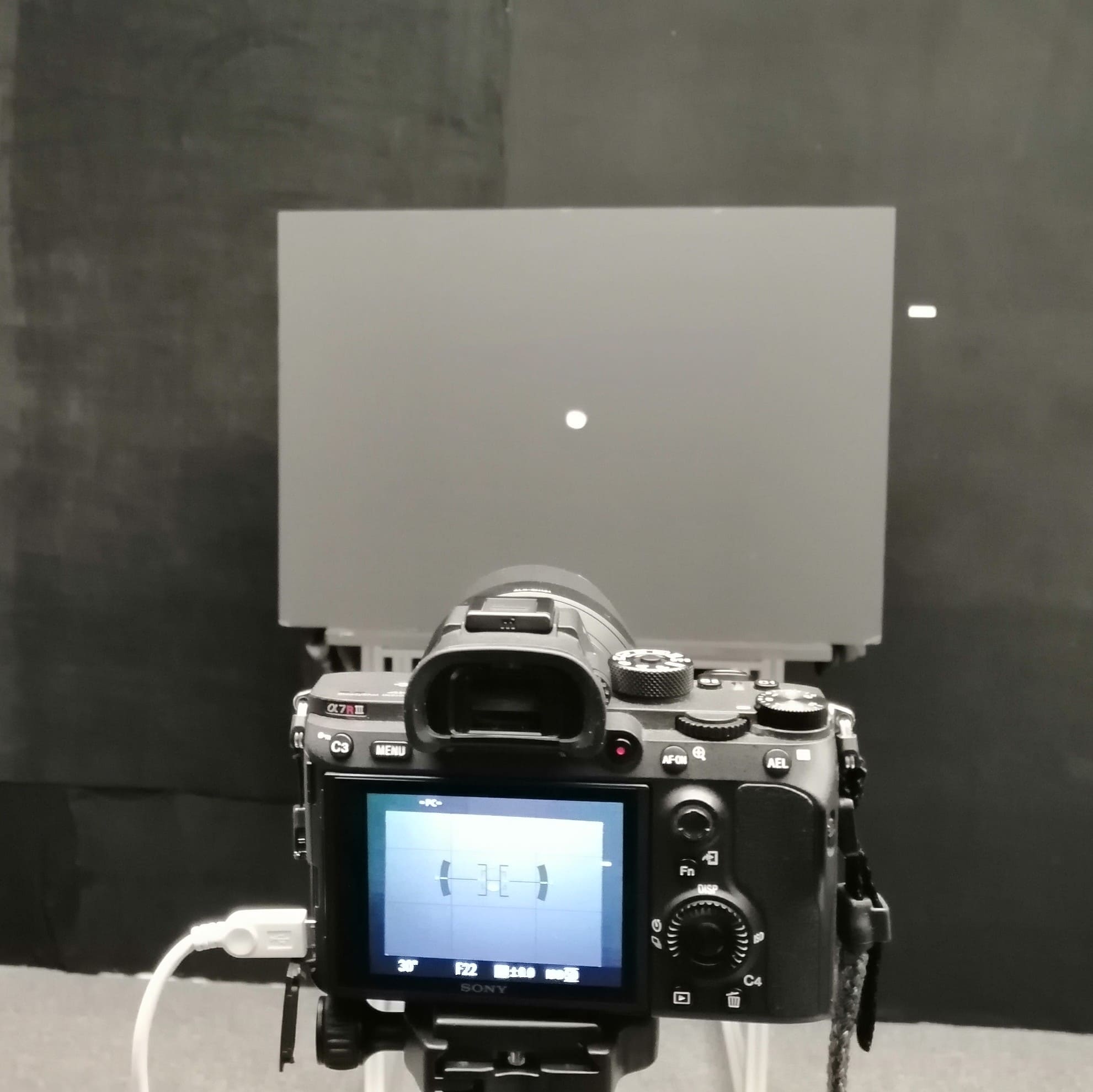}
\caption{Experimental setup to capture high dynamic range image for LSF estimation.}
\label{fig:Saturation-measurement}
\end{figure}

To estimate the LSF, we used an IDS UI-3140CP-M-GL camera with a Navitar MVL4WA wide-angle lens having a focal length of 3.5 mm. The LSF measurements were obtained by capturing High Dynamic Range (HDR) images using a small light source partially obscured by a cardboard cover, leaving a 2 mm circular aperture at the center, as depicted in Figure~\ref{fig:Saturation-measurement}.
We, employed a multi-exposure HDR imaging technique with images taken at varying exposure times and merged using \emph{pfstools}\footnote{http://pfstools.sourceforge.net/}. Exposure times were carefully adjusted to avoid pixel saturation in the shortest exposure image, ensuring optimal image quality.\\
Typically, scattering and reflections of light within the camera lens and its body introduce an additional source of light that indirectly reaches the sensor. Each sensor location receives a small amount of light that does not reach other locations. To simulate this physical phenomenon, it is necessary to represent it in terms of linear radiance values. In many cameras, it can be approximated through a spatially invariant convolution using an LSF. The radiance received by the sensor, denoted as \( l_s \), can be described by:

\begin{equation}
    l_s(x,y) = l_{in}(x,y) \circledast i(x,y),
\end{equation}

where \( l_{in} \) is the incoming radiance projected at pixel coordinates \((x, y)\), and \( i(x,y) \) represents the LSF. To ensure an accurate simulation, the original radiance map $l_{in}$, is first expanded to twice its original resolution and then padded with zeros. This process ensures consistency between the image size and the convolution kernel, which is double the image dimensions. Padding with zeros effectively mimics the behavior of a lens hood, blocking stray light that cannot be focused on the sensor. However, performing convolution with such large kernels directly is computationally expensive. Therefore, in practice, the convolution is typically performed in the Fourier domain:\\ 

\begin{equation}
    L_s(\omega,\phi) = L_{in}(\omega,\phi) \cdot I(\omega,\phi),
\end{equation}
where \( L_s \), \( L_{in} \), and \( I \) denote the Fourier transforms of \( l_s \), \( l_{in} \), and \( i \) respectively.\\
Given the ground truth image \( l_{in} \) (essentially a black field with a central circular light source), the LSF  can theoretically be derived through deconvolution. However, due to noise sensitivity and unreliability, an alternative parameterized approach is used. The LSF is estimated by optimizing the parameters of a radially symmetric function:
\begin{equation}
    i(r) = p_1 \, \delta(r) + p_2 \exp\left( -p_3 \, r^{p_4} \right),
    \label{eq:param-gsf}
\end{equation}
where \( \delta(\cdot) \) is the Dirac delta function, \( p_1, p_2, p_3, p_4 \) are parameters to be estimated, and \( r = \sqrt{x^2 + y^2} \) denotes the radial distance in pixel units. These parameters are optimized by minimizing the following objective function:
\begin{equation}
 \operatorname*{argmin} _ {p_1,p_2,p_3,p_4} \left\| \log(l_{s}) - \log(l_{{capt}}) \right\|_2,
\end{equation}
where \( l_{capt} \) is the captured HDR image, and \( l_{s} \) is the result of convolving the ground truth radiance map with the parametric LSF. The logarithmic scale ensures proper handling of high and low-intensity values of HDR images.
\section{Experimental Results}
In this section, we present both quantitative and qualitative comparisons of the proposed method against state-of-the-art
\begin{figure*}[!h]
\centering
\begin{subfigure}{1.3in}
{\includegraphics[width=1.3in,height=1.3in,keepaspectratio]{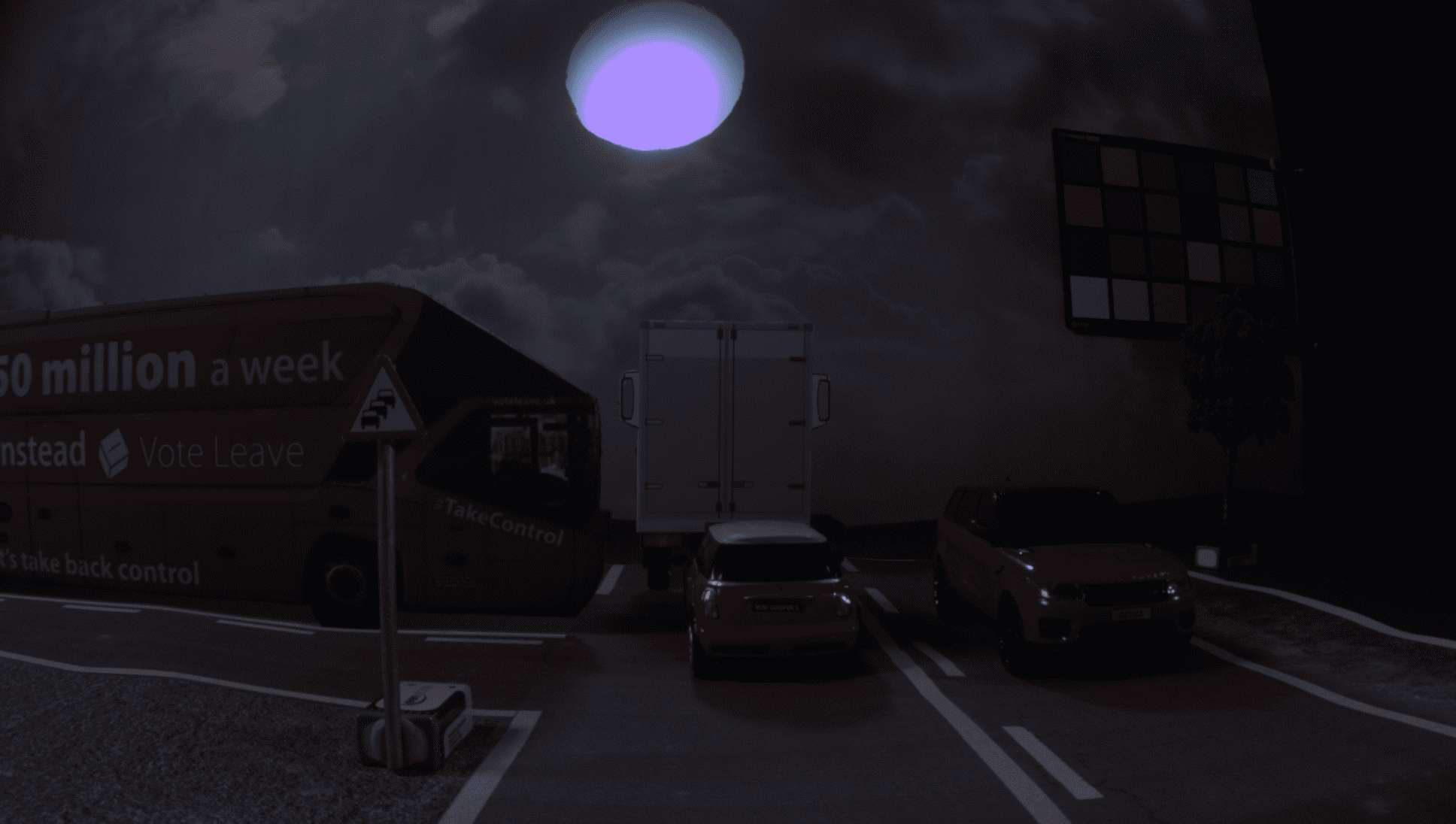}}
 \caption{4.8 [ms]}
\end{subfigure}\hfill
\begin{subfigure}{1.3in}
{\includegraphics[width=1.3in,height=1.3in,keepaspectratio]{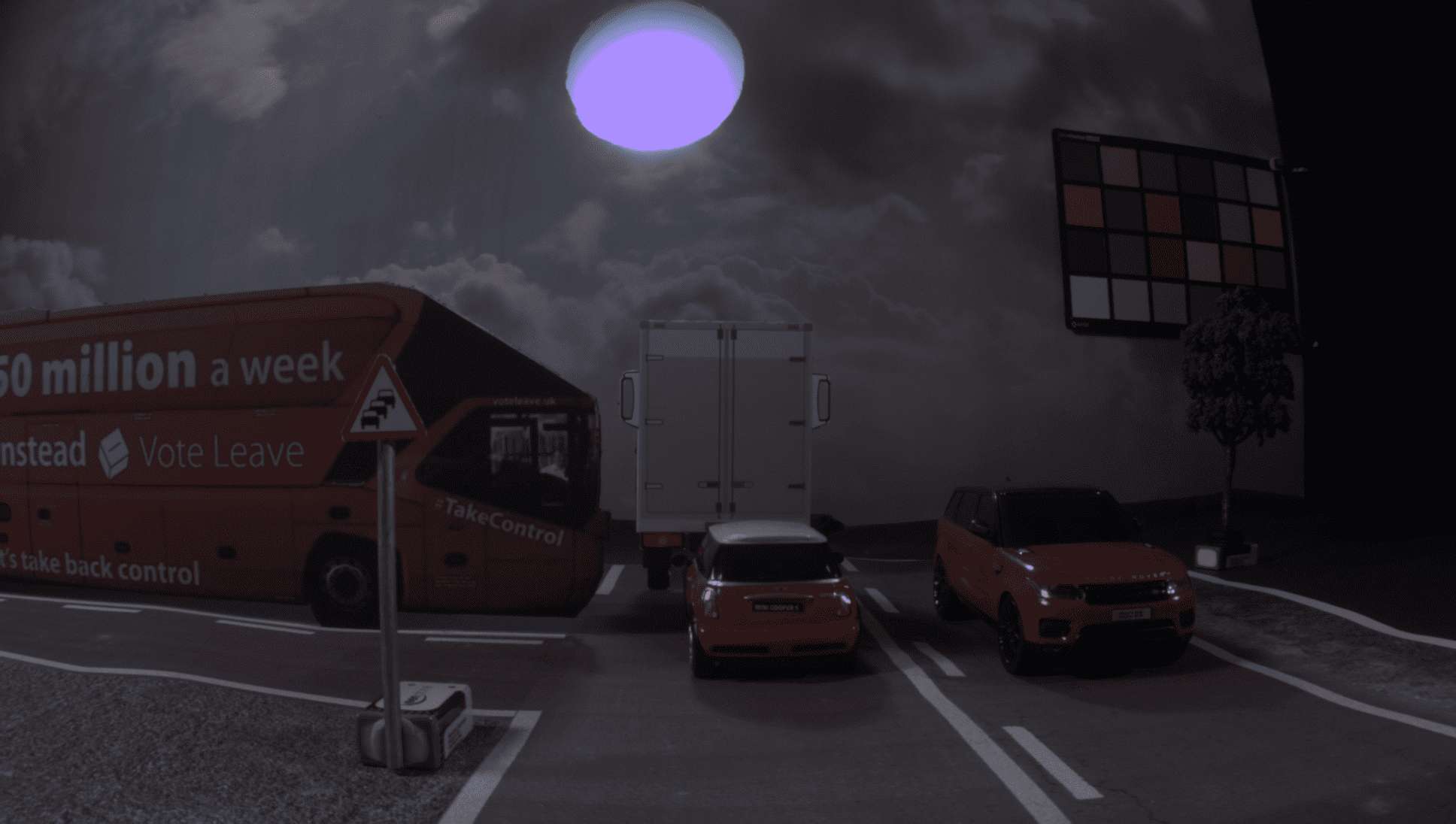}}
 \caption{9.7 [ms]}
\end{subfigure}\hfill
\begin{subfigure}{1.3in}
{\includegraphics[width=1.3in,height=1.3in,keepaspectratio]{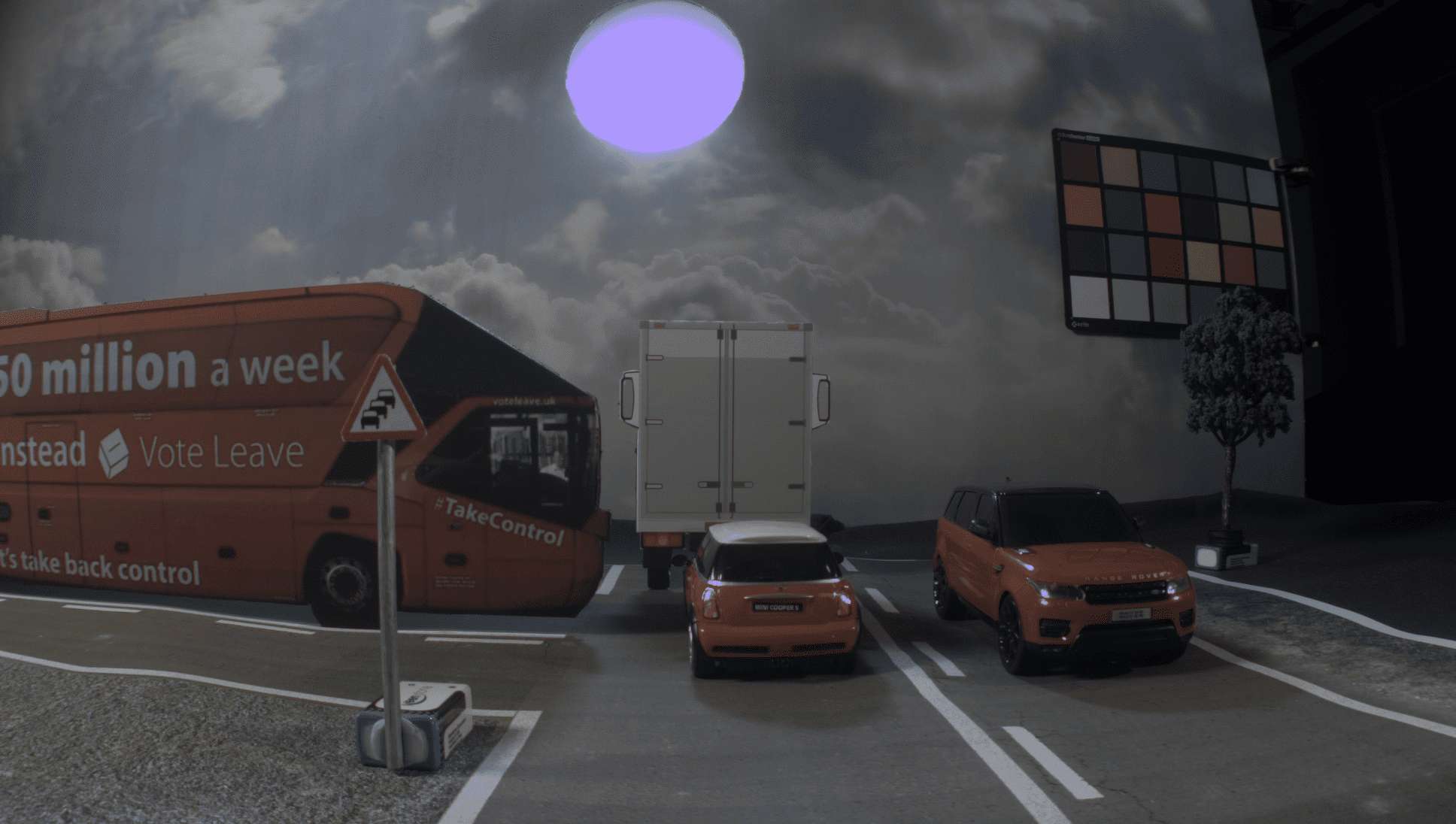}}
 \caption{19.5 [ms]}
\end{subfigure}\hfill
\begin{subfigure}{1.3in}
{\includegraphics[width=1.3in,height=1.3in,keepaspectratio]{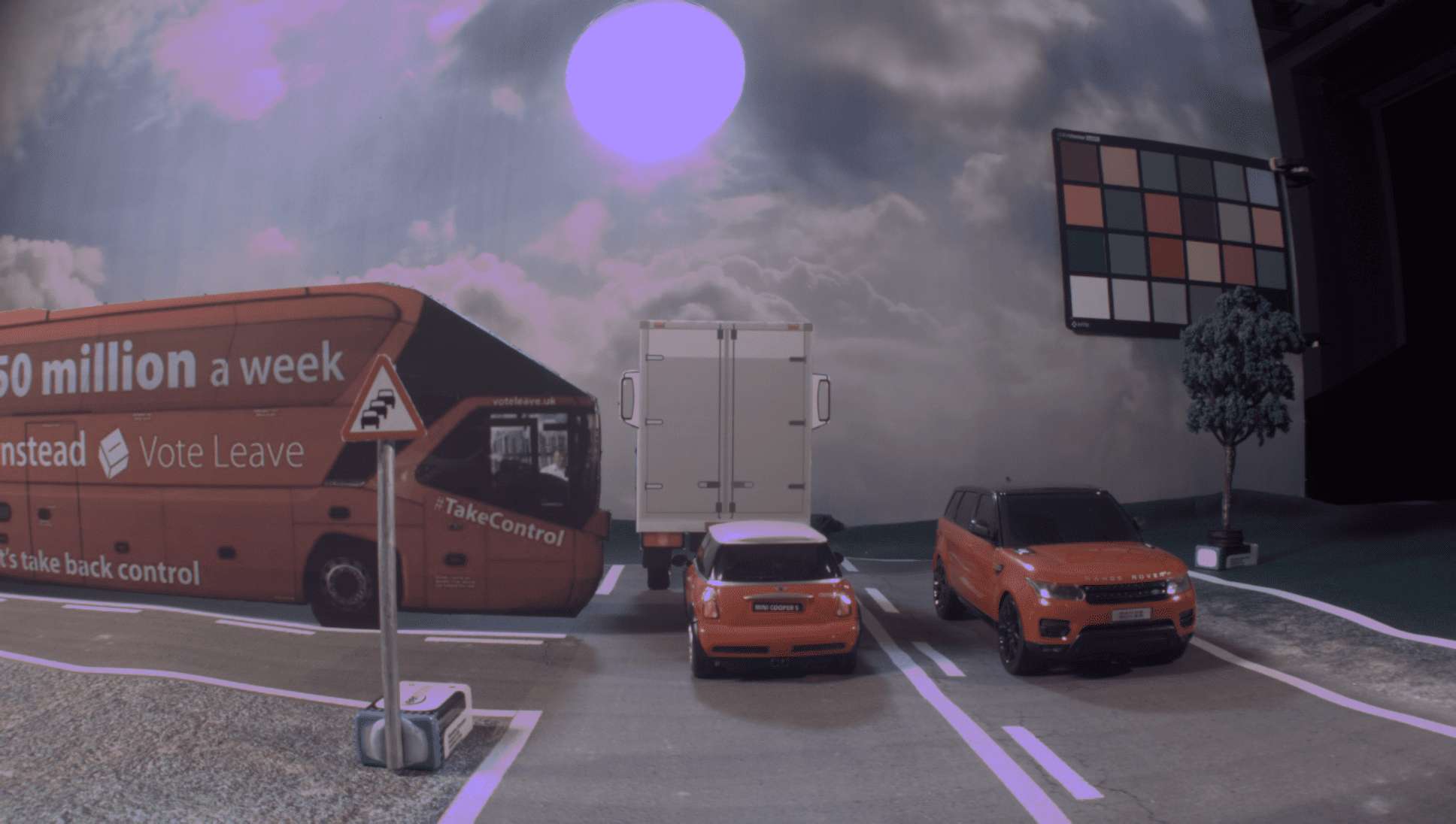}}
 \caption{39.0 [ms]}
\end{subfigure}\hfill
\begin{subfigure}{1.3in}
{\includegraphics[width=1.3in,height=1.3in,keepaspectratio]{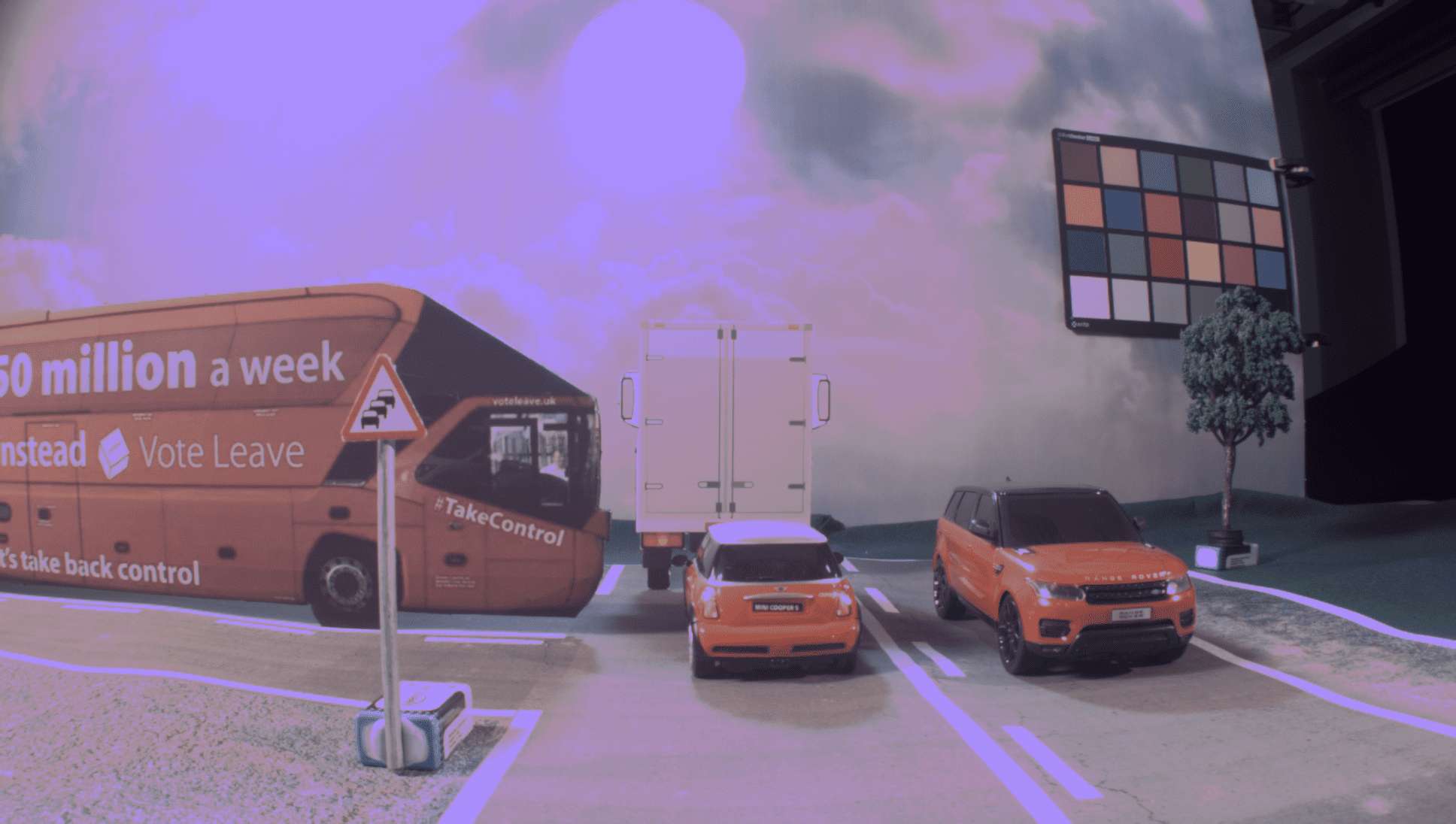}}
 \caption{78.1 [ms]}
\end{subfigure}\hfill\\
\caption{A stack of varying exposure frames for a single image from the tunnel scene in \cite{hanji2021hdr4cv}, utilized to generate the results presented in Figure \ref{Varying_exposure}.}
\label{expourse_view}
\end{figure*}
image deblurring techniques designed specifically for saturation handling, as well as more general deblurring methods. Our approach was evaluated on both synthetic and real datasets to comprehensively demonstrate its robustness. Our deblurring framework operates in two phases: the first phase involves saturation estimation, followed by image deblurring with identified saturated regions in the second phase.\\
In all experiments, we utilized \cite{xu2010two} as the primary blind image deblurring method. Additionally, we tested some other existing deblurring techniques to ensure that the proposed framework can be adapted for integration with the latest blind image deblurring methods. This adaptability is intended to leverage advancements in image deblurring research for improved performance and applicability\\
{To evaluate the computational efficiency of the proposed method, we conducted runtime comparisons using a workstation equipped with an Intel® Core™ Ultra 7 155H CPU (1.40 GHz), 16 GB of RAM (5600 MHz), running a 64-bit Windows OS. We compared the total execution time of our framework against a subset of representative learning-based and non-learning-based deblurring methods, as summarized in Table I. For the learning-based methods (e.g., Liang et al. \cite{chen2021learning}  and Mao et al. \cite{LoFormer}), we measured only the inference time to ensure fair comparison. Our results indicate that the proposed method, particularly when integrated with Xu et al's \cite{xu2010two} deblurring approach, achieves the best overall performance in terms of runtime, with a total average processing time.\\
\begin{table}[]
\begin{tabular}{|l|l|l|l|l|}
\hline
Method                                                                                                                          & Method type  & Image Size & \begin{tabular}[c]{@{}l@{}}Sat. Est.\\  time.\end{tabular} & Total time \\ \hline
\begin{tabular}[c]{@{}l@{}}Proposed W/Xu\\  \emph{et al.} \cite{xu2010two} \end{tabular}       & Non-learning & 1280 x720  & 83s                                                        & 126s            \\ \hline
\begin{tabular}[c]{@{}l@{}}Proposed W/\\  Wen \emph{et al.} \cite{wen2020simple}\end{tabular} & Non-learning & 1280 x720  & 83s                                                        & 1586s           \\ \hline
Hu \cite{hu2014deblurring}                                                                                     & Non-learning & 1280 x720  & --                                                         & 143s            \\ \hline
chen \cite{chen2021blind}                                                                                      & Non-learning & 1280 x720  & --                                                         & 124s            \\ \hline
Liang \cite{chen2021learning}                                                                                  & Learning     & 1280 x720  & --                                                         & 318s            \\ \hline
Mao \cite{LoFormer}                                                                                            & Learning     & 1280 x720  & --                                                         & 247s            \\ \hline
\end{tabular}
\caption{Comparison of the computational efficiency of the proposed method. The reported time values represent the average total computation time measured over 10 test images.}
\label{Comp_time}
\end{table}
{We evaluated how different LSF estimates derived from multiple camera-lens systems affect overall deblurring performance. Specifically, we estimated LSF parameters 
p1,p2,p3,p4 using HDR images captured with various camera and lens combinations. 
The resulting LSFs were used independently in our deblurring framework on a common test dataset \cite{hanji2021hdr4cv}. As shown in Figure \ref{ablation}, despite the variation in camera/lens system, the variation in the performance of the proposed deblurring framework is insignificant. This demonstrates the adaptability of the framework to LSF variability and confirms that the parametric formulation is not overly sensitive to camera-specific nuances. Furthermore, to underscore the value of our parametric LSF estimation approach, we also include performance using a deconvolution-based non-parametric LSF estimation. The results demonstrate that in our case, the parametric model with optimized parameters offers both physical plausibility and practical effectiveness.}
\begin{figure}[!h]
\centering
{\includegraphics[trim={45 7 45 23},clip,width=3.5in,height=3.5in,keepaspectratio]{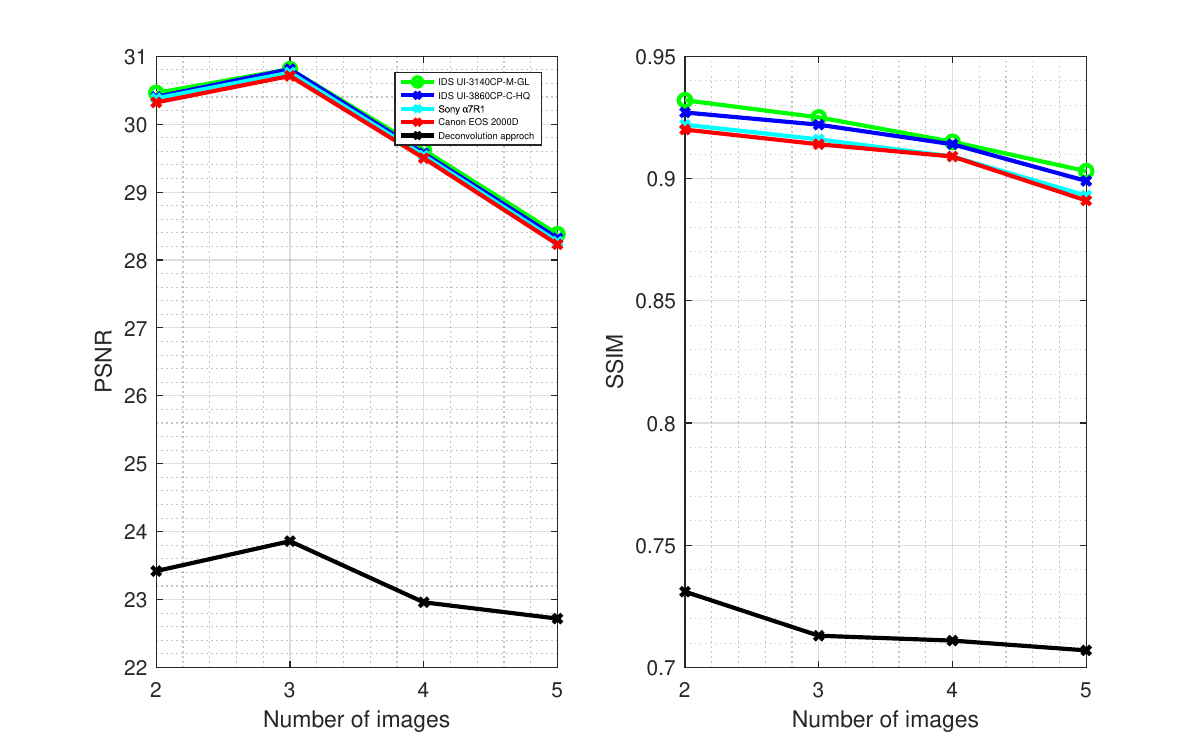}}
\caption{{Impact of LSF from different camera and lens combinations, and parametric vs non-parametric approach to LSF estimation on the performance of the proposed deblurring framework.}}
\label{ablation}
\end{figure}\\
\color{black}To rigorously evaluate the robustness of the proposed method under varying levels of brightness, we utilized the dataset presented in \cite{hanji2021hdr4cv}, which offers a collection of images with varying exposure levels. Specifically, we focused on the tunnel scene due to its high dynamic range characteristics, as illustrated in Figure \ref{expourse_view}. This dataset comprises a sequence of 100 images capturing a static scene with controlled camera motion. To generate the blurred input from the sequence of images, we merged five consecutive frames for each exposure level to produce the results shown in Figure \ref{Varying_exposure}.
\begin{figure}[!h]
\centering
{\includegraphics[trim={45 7 46 23},clip,width=3.5in,height=3.5in,keepaspectratio]{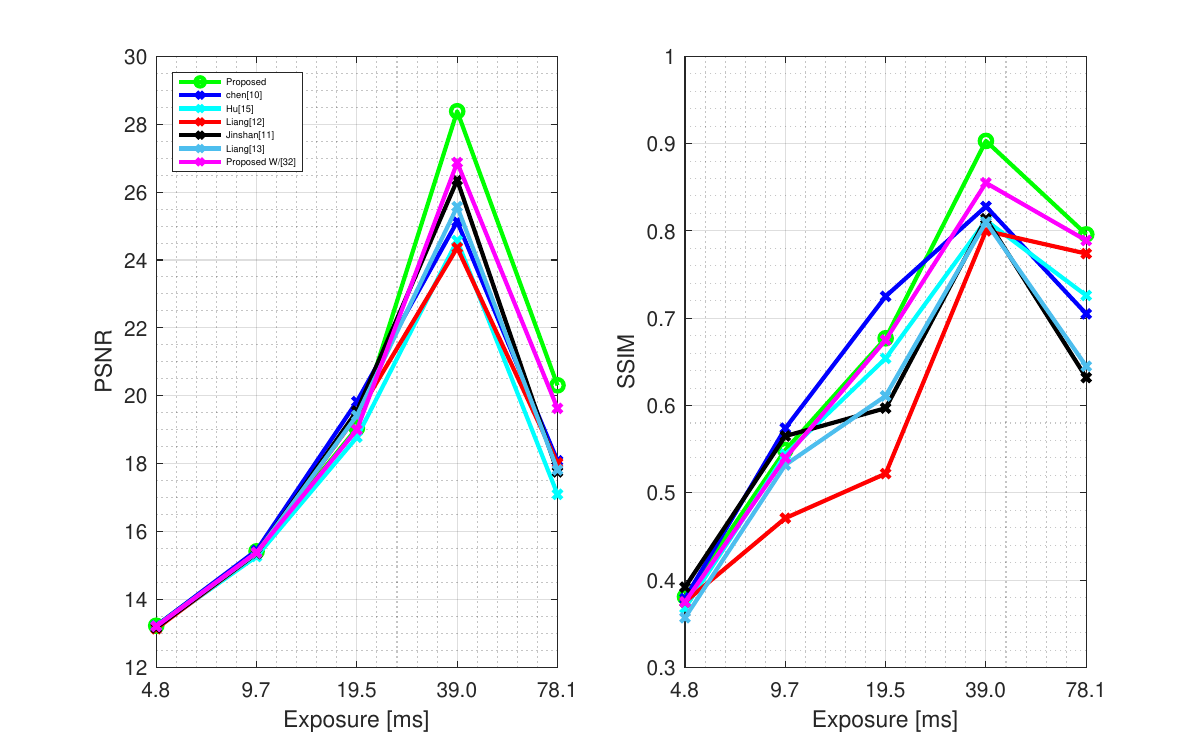}}
\caption{Performance comparison of the proposed method with existing saturation-aware image deblurring techniques, evaluated across varying exposure levels.\color{black}}
\label{Varying_exposure}
\end{figure}\\
In this experiment, we analyze the comparative performance of various saturation-aware deblurring methods \cite{restormer}, \cite{zhang2017learning} against the proposed approach under diverse exposure conditions. The proposed framework is built upon an underlying deconvolution deblurring method \cite{8451604}, enhanced with saturation estimation. At very low exposures, saturation is minimal and thus the saturation-aware component becomes less influential. Consequently, the performance of our method converges to that of the base deblurring algorithm used within the framework, resulting in comparable performance to other methods. In contrast, at moderate to high exposures where saturation is more pronounced, our method clearly outperforms others by accurately estimating radiance in saturated regions \cite{10507029}.\\ 
In Figure \ref{varying_blur}, we provide a quantitative evaluation \cite{ 11259078} of the proposed method,
\begin{figure}[!h]
\centering
{\includegraphics[trim={45 7 45 23},clip,width=3.5in,height=3.5in,keepaspectratio]{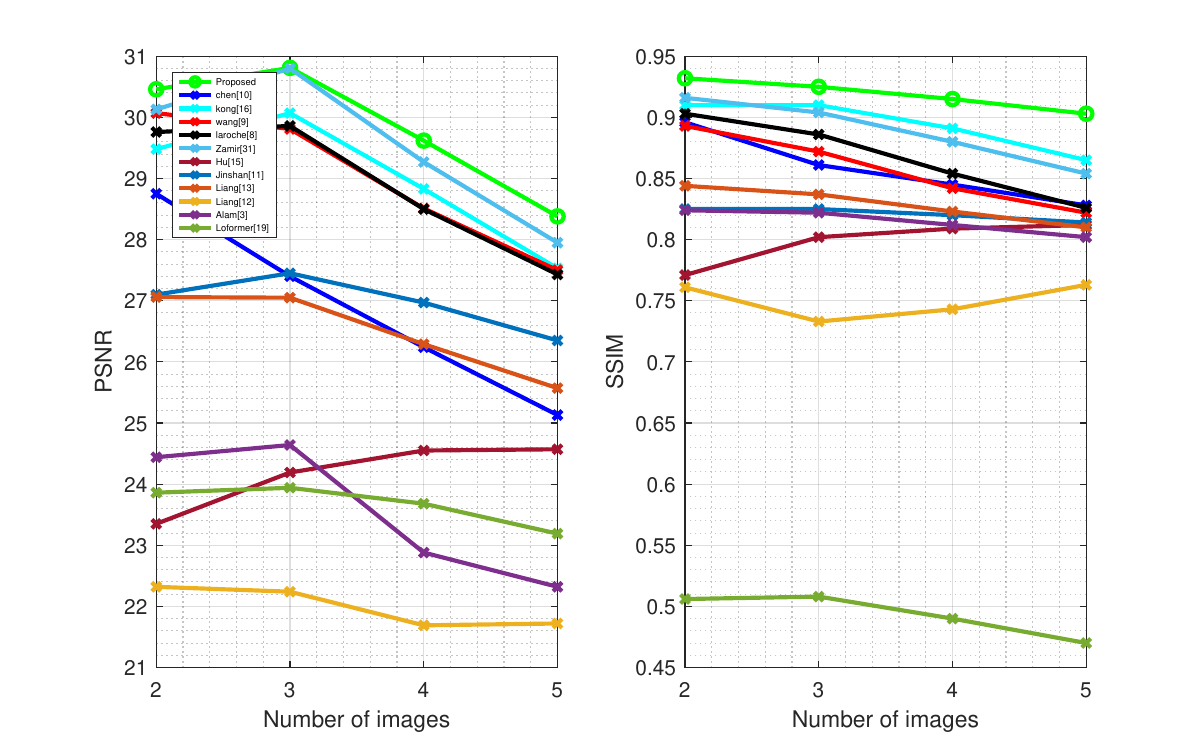}}
\caption{Comparison of state-of-the-art saturation-aware and general-purpose deblurring methods with the proposed technique on a dataset exhibiting varying levels of blur. A randomly selected sample image from the blur dataset \cite{hanji2021hdr4cv} is shown in Figure \ref{blur_view}.}
\label{varying_blur}
\end{figure}
\begin{figure}[!h]
\centering
\begin{subfigure}{1.70in}
{\includegraphics[width=1.70in,height=1.70in,keepaspectratio]{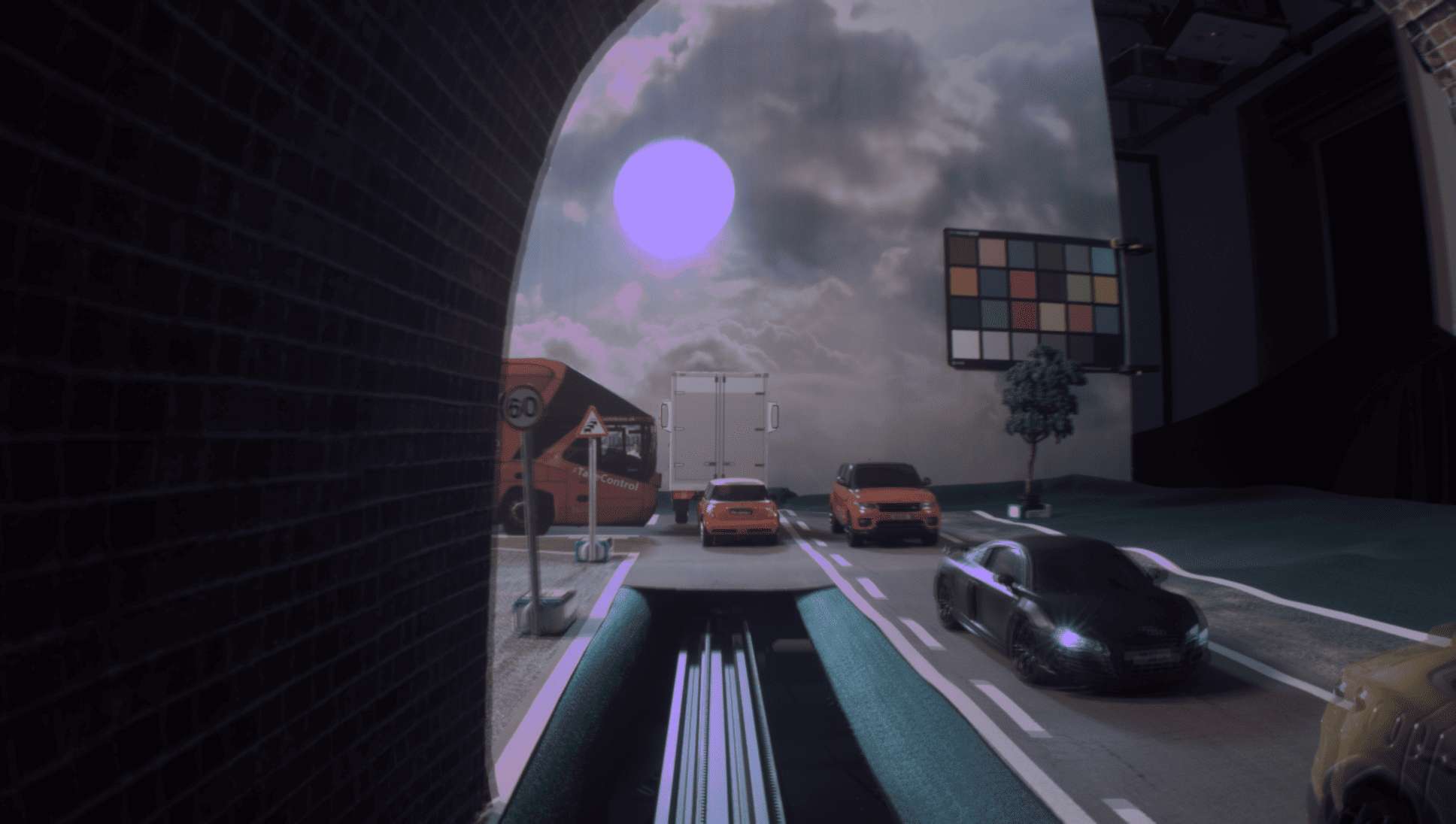}}
 \caption{Im $\times$ 2}
\end{subfigure}
\begin{subfigure}{1.70in}
{\includegraphics[width=1.70in,height=1.70in,keepaspectratio]{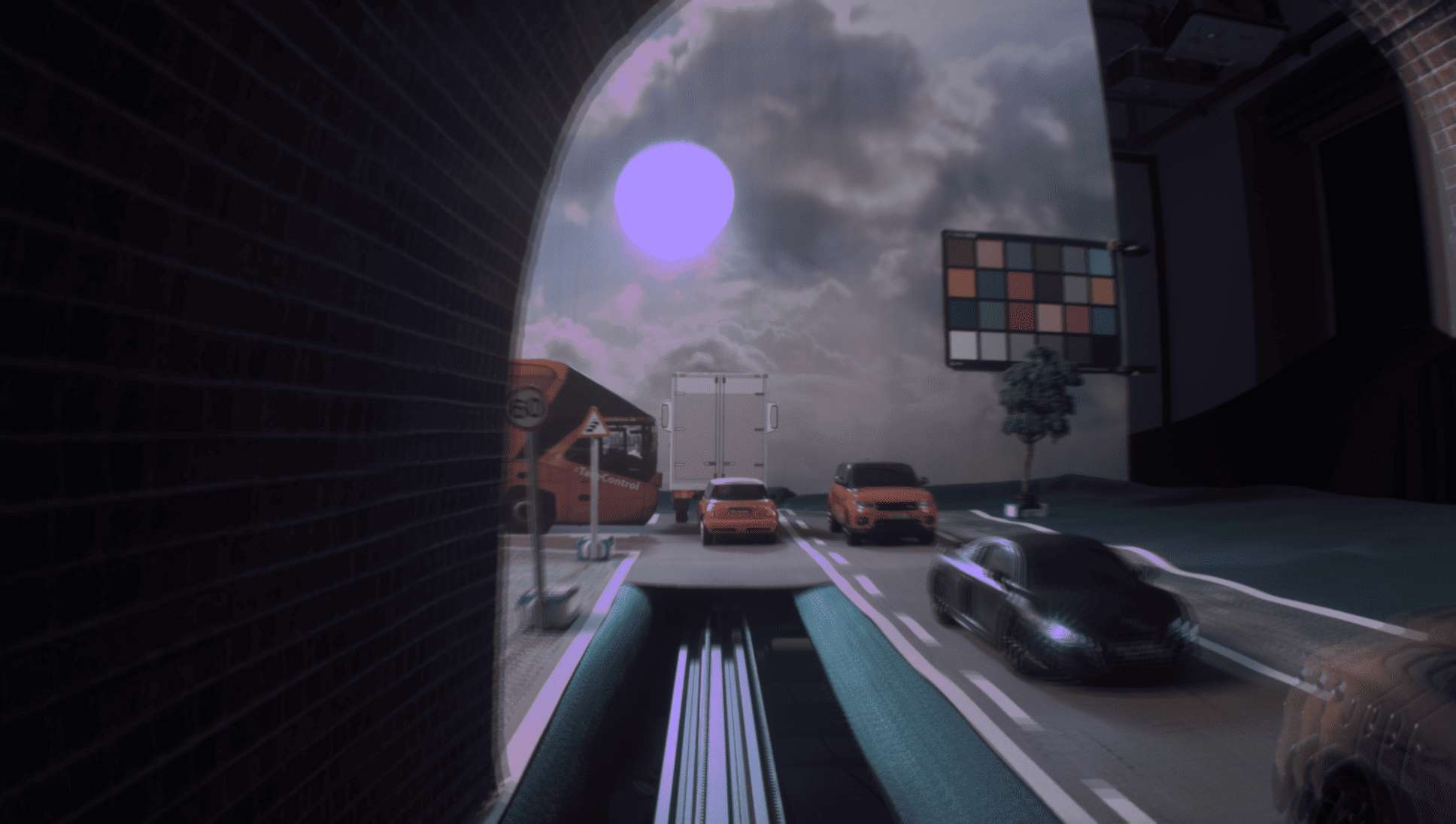}}
 \caption{Im $\times$ 3}
\end{subfigure}\\
\begin{subfigure}{1.70in}
{\includegraphics[width=1.70in,height=1.70in,keepaspectratio]{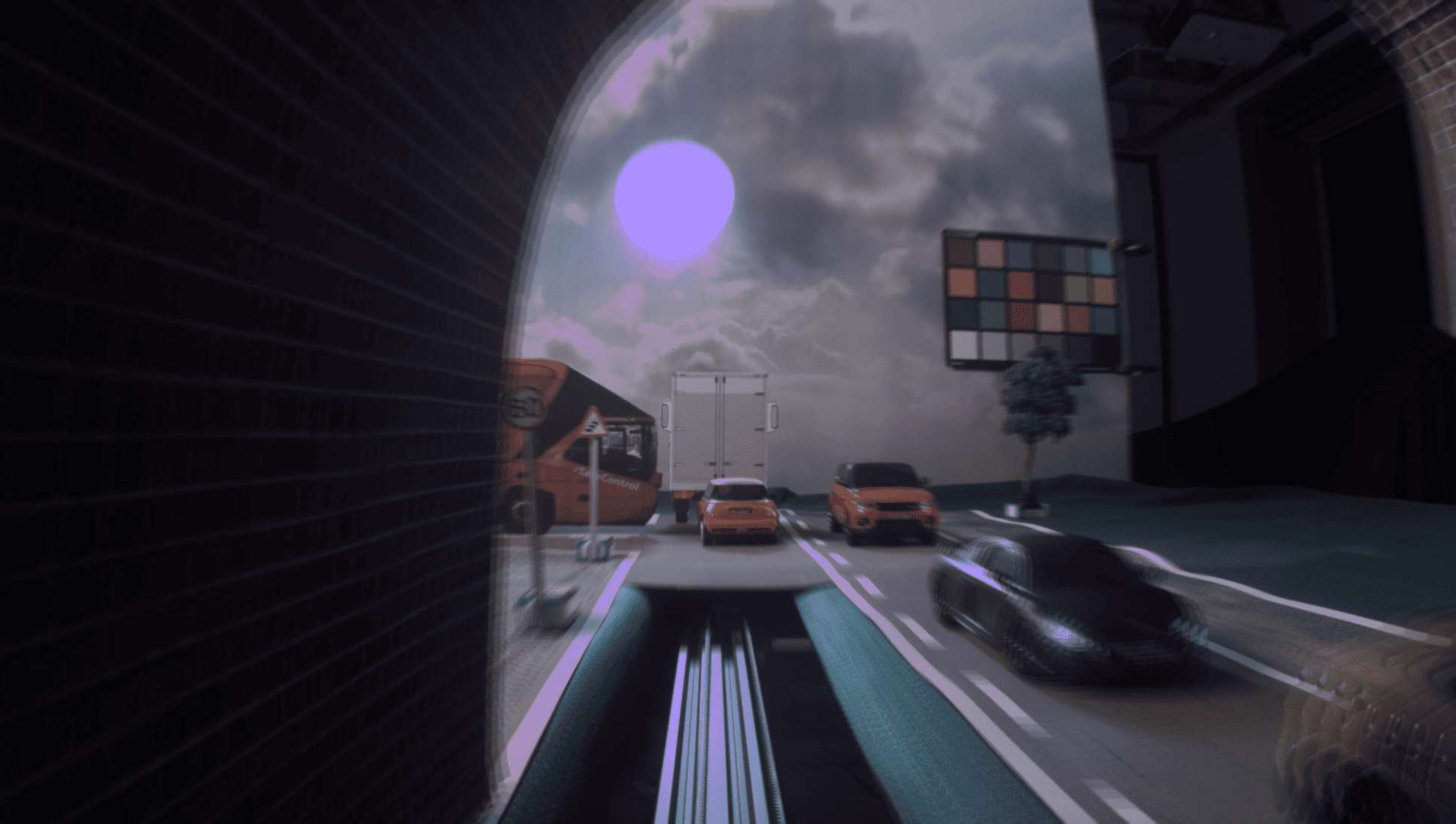}}
 \caption{Im $\times$ 4}
\end{subfigure}
\begin{subfigure}{1.70in}
{\includegraphics[width=1.70in,height=1.70in,keepaspectratio]{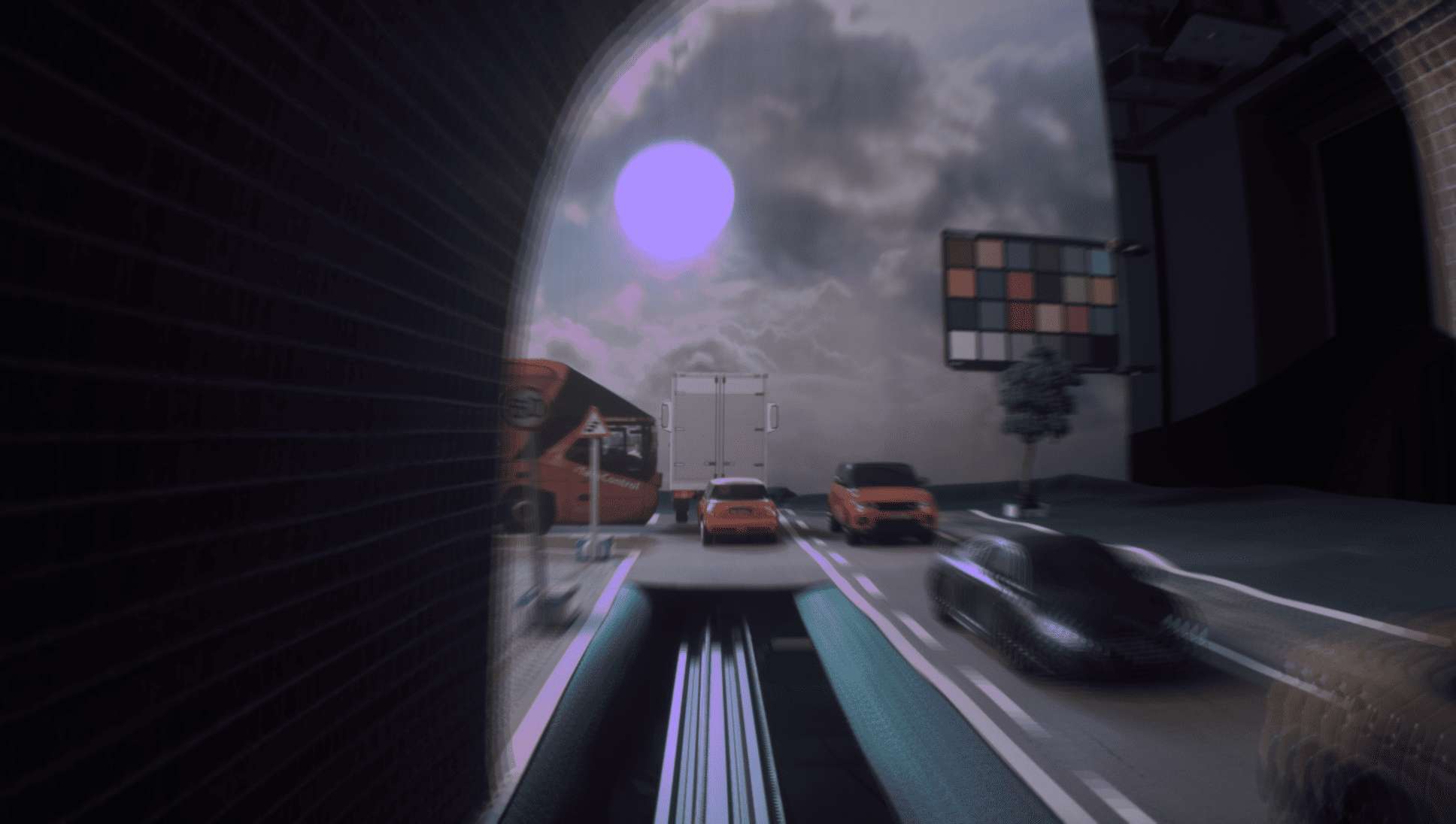}}
 \caption{Im $\times$ 5}
\end{subfigure}
\caption{A Sample representative image illustrating varying levels of blur from the dataset used to generate the results shown in Figure \ref{varying_blur}. }
\label{blur_view}
\end{figure}
\begin{table*}[!h]
\begin{tabular}{|l|l|llllll|llllll|}
\hline
\multirow{2}{*}{Method} & \multirow{2}{*}{Saturation handling} & \multicolumn{6}{l|}{Geometric Mean}                                                                                               & \multicolumn{6}{l|}{SSIM-Weighted PSNR}                                                                                   \\ \cline{3-14} 
                        &                                      & \multicolumn{1}{c|}{Im2}   & \multicolumn{1}{c|}{Im3}   & \multicolumn{1}{c|}{Im4}   & \multicolumn{2}{c|}{Im5}   & \multicolumn{1}{c|}{Avg.} & \multicolumn{1}{c|}{Im2}   & \multicolumn{1}{c|}{Im3}   & \multicolumn{1}{c|}{Im4}   & \multicolumn{2}{c|}{Im5}   & Avg.  \\ \hline
Proposed                & Saturation Handling                  & \multicolumn{1}{l|}{5.32} & \multicolumn{1}{l|}{5.33} & \multicolumn{1}{l|}{5.20} & \multicolumn{2}{l|}{5.06} & 5.23                     & \multicolumn{1}{l|}{28.38} & \multicolumn{1}{l|}{28.49} & \multicolumn{1}{l|}{27.11} & \multicolumn{2}{l|}{25.63} & 27.40 \\ \hline
chen\cite{chen2021blind}            & Saturation Handling                  & \multicolumn{1}{l|}{5.07} & \multicolumn{1}{l|}{4.85} & \multicolumn{1}{l|}{4.70} & \multicolumn{2}{l|}{4.56} & 4.80                     & \multicolumn{1}{l|}{25.77} & \multicolumn{1}{l|}{23.60} & \multicolumn{1}{l|}{22.17} & \multicolumn{2}{l|}{20.81} & 23.08 \\ \hline
Kong \cite{kong2023efficient}            & General                              & \multicolumn{1}{l|}{5.17} & \multicolumn{1}{l|}{5.23} & \multicolumn{1}{l|}{5.06} & \multicolumn{2}{l|}{4.87} & 5.08                     & \multicolumn{1}{l|}{26.83} & \multicolumn{1}{l|}{27.36} & \multicolumn{1}{l|}{25.70} & \multicolumn{2}{l|}{23.80} & 25.92 \\ \hline
Wang \cite{wang2023ddnm}             & Night                                & \multicolumn{1}{l|}{5.18} & \multicolumn{1}{l|}{5.09} & \multicolumn{1}{l|}{4.89} & \multicolumn{2}{l|}{4.75} & 4.98                     & \multicolumn{1}{l|}{26.85} & \multicolumn{1}{l|}{26.00} & \multicolumn{1}{l|}{24.01} & \multicolumn{2}{l|}{22.61} & 24.87 \\ \hline
Laroche \cite{laroche2024fast}          & Saturation Handling                  & \multicolumn{1}{l|}{5.18} & \multicolumn{1}{l|}{5.14} & \multicolumn{1}{l|}{4.93} & \multicolumn{2}{l|}{4.75} & 5.00                     & \multicolumn{1}{l|}{26.87} & \multicolumn{1}{l|}{26.46} & \multicolumn{1}{l|}{24.36} & \multicolumn{2}{l|}{22.66} & 25.09 \\ \hline
Hu \cite{hu2014deblurring}              & Saturation Handling                  & \multicolumn{1}{l|}{4.24} & \multicolumn{1}{l|}{4.40} & \multicolumn{1}{l|}{4.45} & \multicolumn{2}{l|}{4.46} & 4.39                     & \multicolumn{1}{l|}{17.99} & \multicolumn{1}{l|}{19.40} & \multicolumn{1}{l|}{19.87} & \multicolumn{2}{l|}{19.94} & 19.30 \\ \hline
Jinshan \cite{pan2016blind}         & Night                                & \multicolumn{1}{l|}{4.72} & \multicolumn{1}{l|}{4.75} & \multicolumn{1}{l|}{4.70} & \multicolumn{2}{l|}{4.63} & 4.70                     & \multicolumn{1}{l|}{22.36} & \multicolumn{1}{l|}{22.66} & \multicolumn{1}{l|}{22.12} & \multicolumn{2}{l|}{21.44} & 22.15 \\ \hline
Liang \cite{chen2020oid}           & Saturation Handling                  & \multicolumn{1}{l|}{4.77} & \multicolumn{1}{l|}{4.75} & \multicolumn{1}{l|}{4.65} & \multicolumn{2}{l|}{4.55} & 4.68                     & \multicolumn{1}{l|}{22.82} & \multicolumn{1}{l|}{22.63} & \multicolumn{1}{l|}{21.62} & \multicolumn{2}{l|}{20.73} & 21.95 \\ \hline
Liang \cite{chen2021learning}           & Saturation Handling                  & \multicolumn{1}{l|}{4.12} & \multicolumn{1}{l|}{4.03} & \multicolumn{1}{l|}{4.01} & \multicolumn{2}{l|}{4.07} & 4.06                     & \multicolumn{1}{l|}{18.33} & \multicolumn{1}{l|}{18.00} & \multicolumn{1}{l|}{17.18} & \multicolumn{2}{l|}{16.16} & 17.42 \\ \hline
Alam \cite{ALAM2019}             & General                              & \multicolumn{1}{l|}{4.48} & \multicolumn{1}{l|}{4.50} & \multicolumn{1}{l|}{4.31} & \multicolumn{2}{l|}{4.23} & 4.38                      & \multicolumn{1}{l|}{19.96} & \multicolumn{1}{l|}{19.62} & \multicolumn{1}{l|}{18.54} & \multicolumn{2}{l|}{17.60} & 18.93 \\ \hline
Zamir \cite{restormer}   & General                              & \multicolumn{1}{l|}{5.25} & \multicolumn{1}{l|}{5.27} & \multicolumn{1}{l|}{5.07} & \multicolumn{2}{l|}{4.88} & 5.12                      & \multicolumn{1}{l|}{27.59} & \multicolumn{1}{l|}{27.84} & \multicolumn{1}{l|}{25.75} & \multicolumn{2}{l|}{23.86} & 26.26 \\ \hline
Mao \cite{LoFormer} & General                              & \multicolumn{1}{l|}{3.47} & \multicolumn{1}{l|}{3.48} & \multicolumn{1}{l|}{3.40} & \multicolumn{2}{l|}{3.30} & 3.41                      & \multicolumn{1}{l|}{12.07} & \multicolumn{1}{l|}{12.16} & \multicolumn{1}{l|}{11.60} & \multicolumn{2}{l|}{10.89} & 11.68 \\ \hline
\end{tabular}
\caption{Comparative evaluation of the proposed method and existing state-of-the-art deblurring techniques using geometric mean and SSIM-Weighted PSNR for holistic assessment considering both signal fidelity and perceptual aspects across varying blur conditions.}
\label{varying_blur_metrics}
\end{table*}
\begin{figure*}[!h]
\centering
\begin{subfigure}{1.57in}
{\includegraphics[width=1.57in,height=1.57in,keepaspectratio]{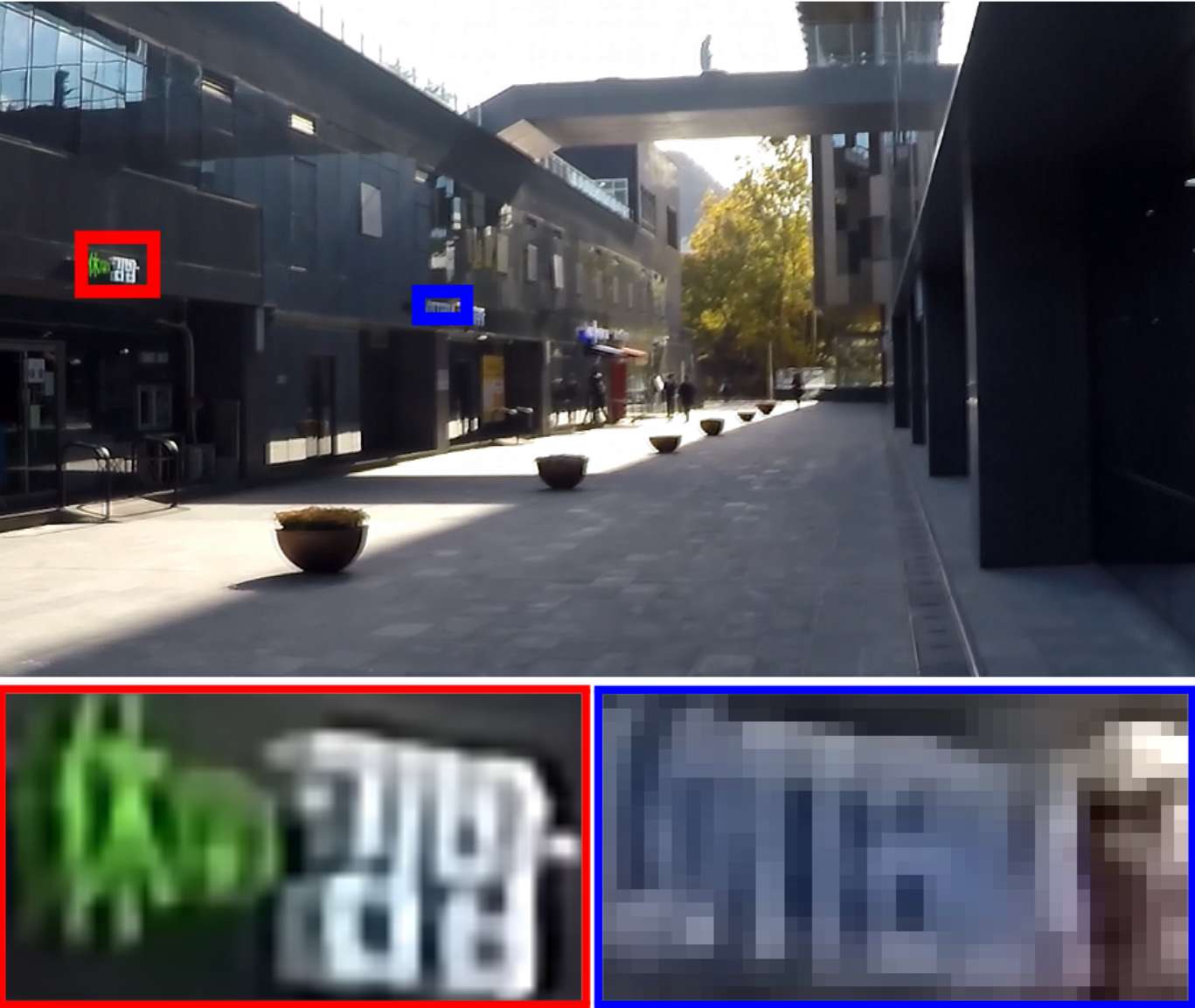}}
 \caption{ Proposed ( Xu \emph{et al.} \cite{xu2010two}) \\ \textbf{PSNR = 25.00, SSIM = 0.89 }}
\end{subfigure}
\begin{subfigure}{1.52in}
{\includegraphics[width=1.52in,height=1.52in,keepaspectratio]{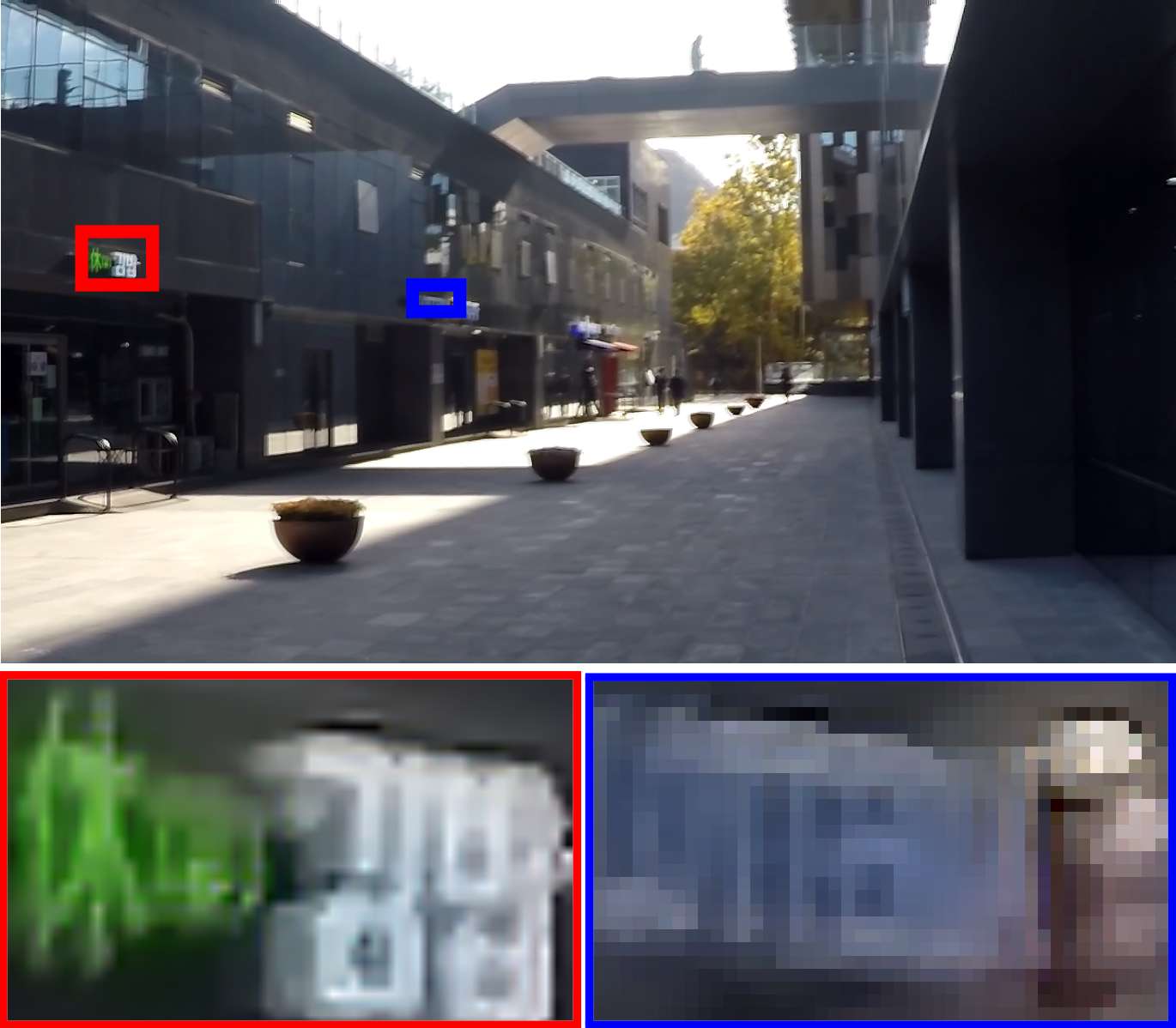}}
 \caption{ Proposed (Wen \emph{et al.} \cite{wen2020simple}) \\ \textbf{PSNR = 24.87, SSIM = 0.88}}
\end{subfigure}
\begin{subfigure}{1.52in}
{\includegraphics[width=1.52in,height=1.52in,keepaspectratio]{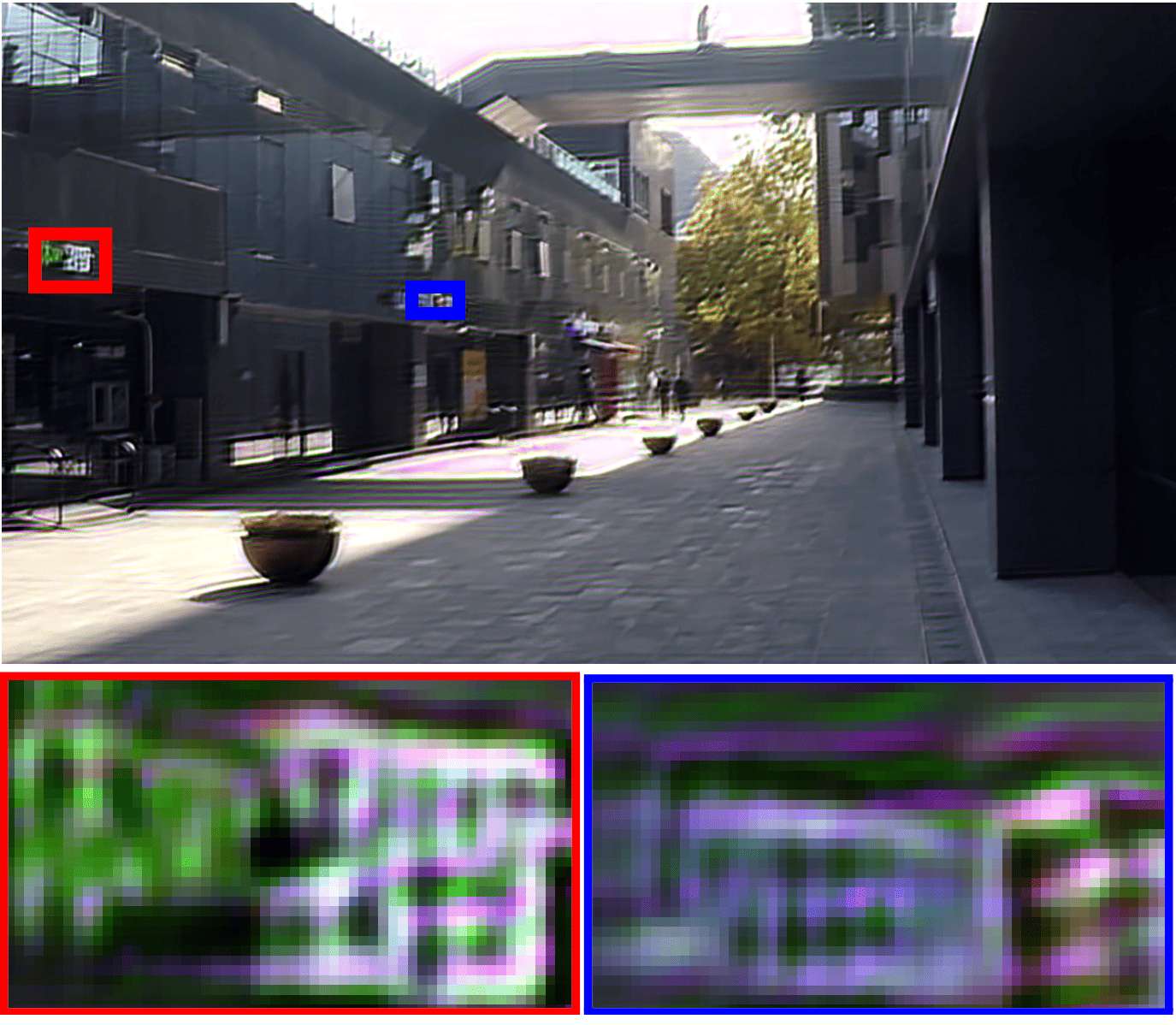}}
 \caption{ chen \emph{et al.} \cite{chen2021learning} \\ \textbf{PSNR = 19.20, SSIM = 0.74 }}
\end{subfigure}
\begin{subfigure}{1.5in}
{\includegraphics[width=1.5in,height=1.5in,keepaspectratio]{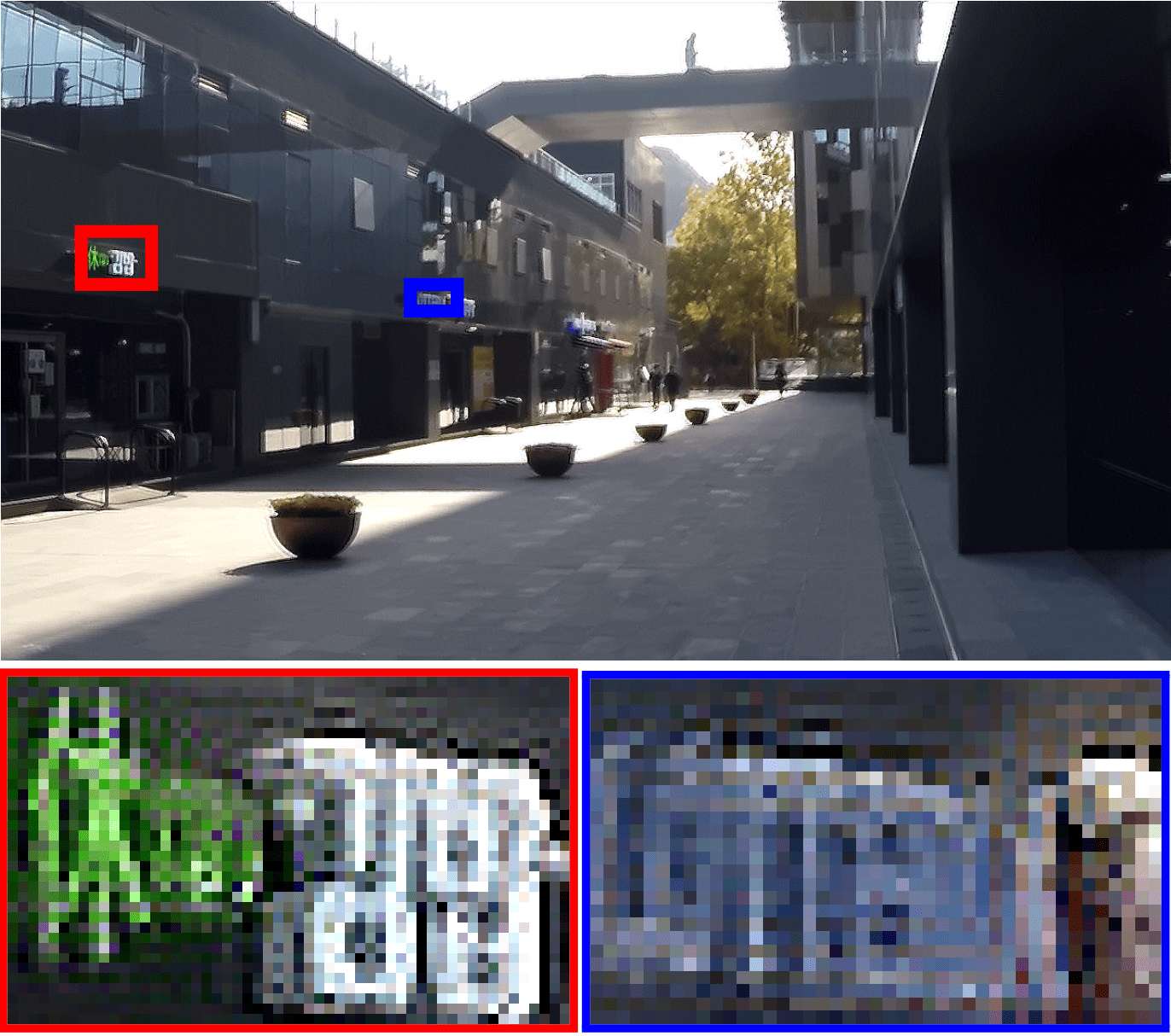}}
 \caption{ Pan \emph{et al.} \cite{pan2016blind} \\ \textbf{PSNR = 22.07, SSIM = 0.77}}
\end{subfigure}\\
\begin{subfigure}{1.5in}
{\includegraphics[width=1.5in,height=1.5in,keepaspectratio]{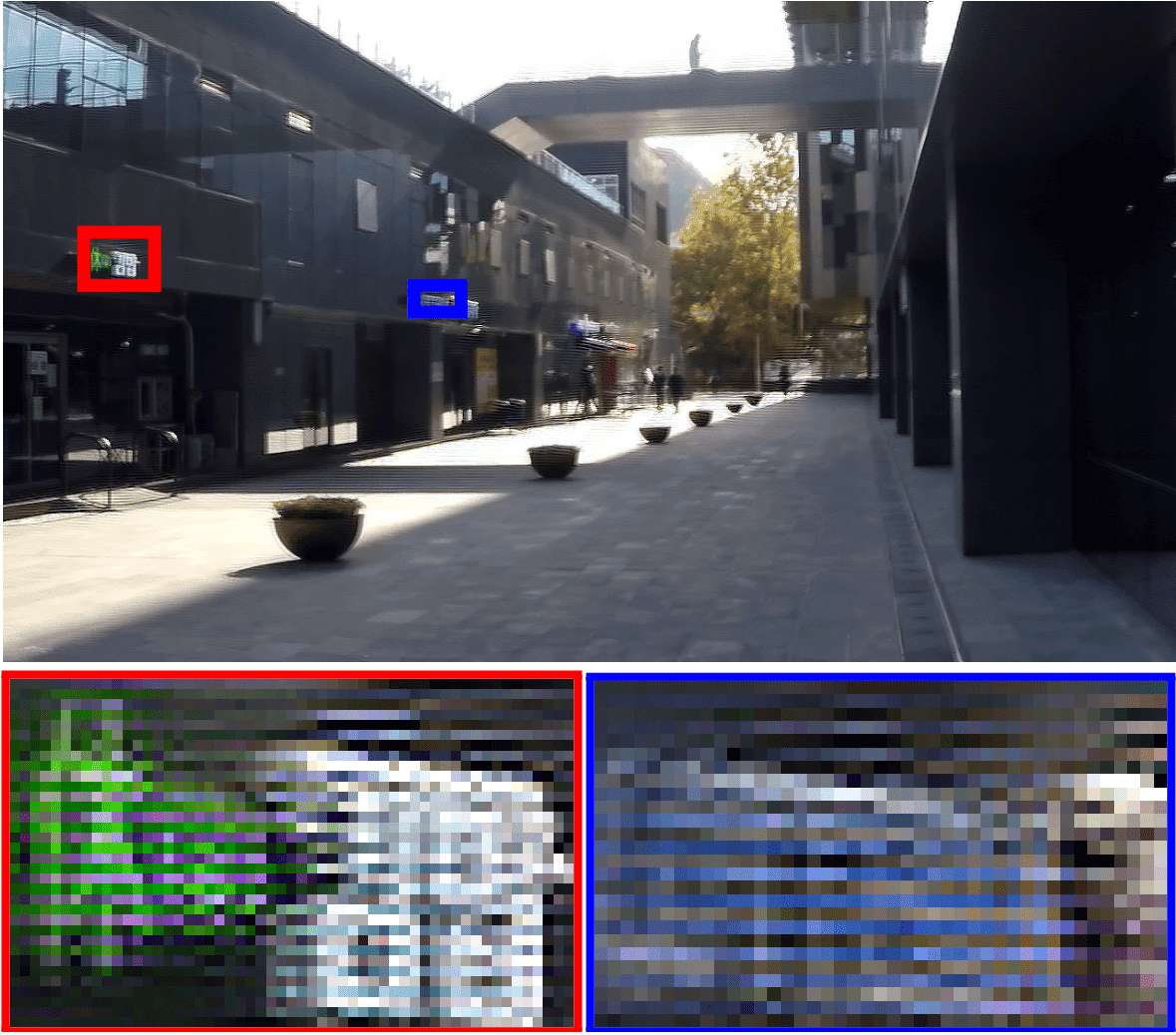}}
 \caption{ Chen \emph{et al.} \cite{chen2020oid} \\ \textbf{PSNR = 22.22, SSIM = 0.76}}
\end{subfigure}
\begin{subfigure}{1.5in}
{\includegraphics[width=1.5in,height=1.5in,keepaspectratio]{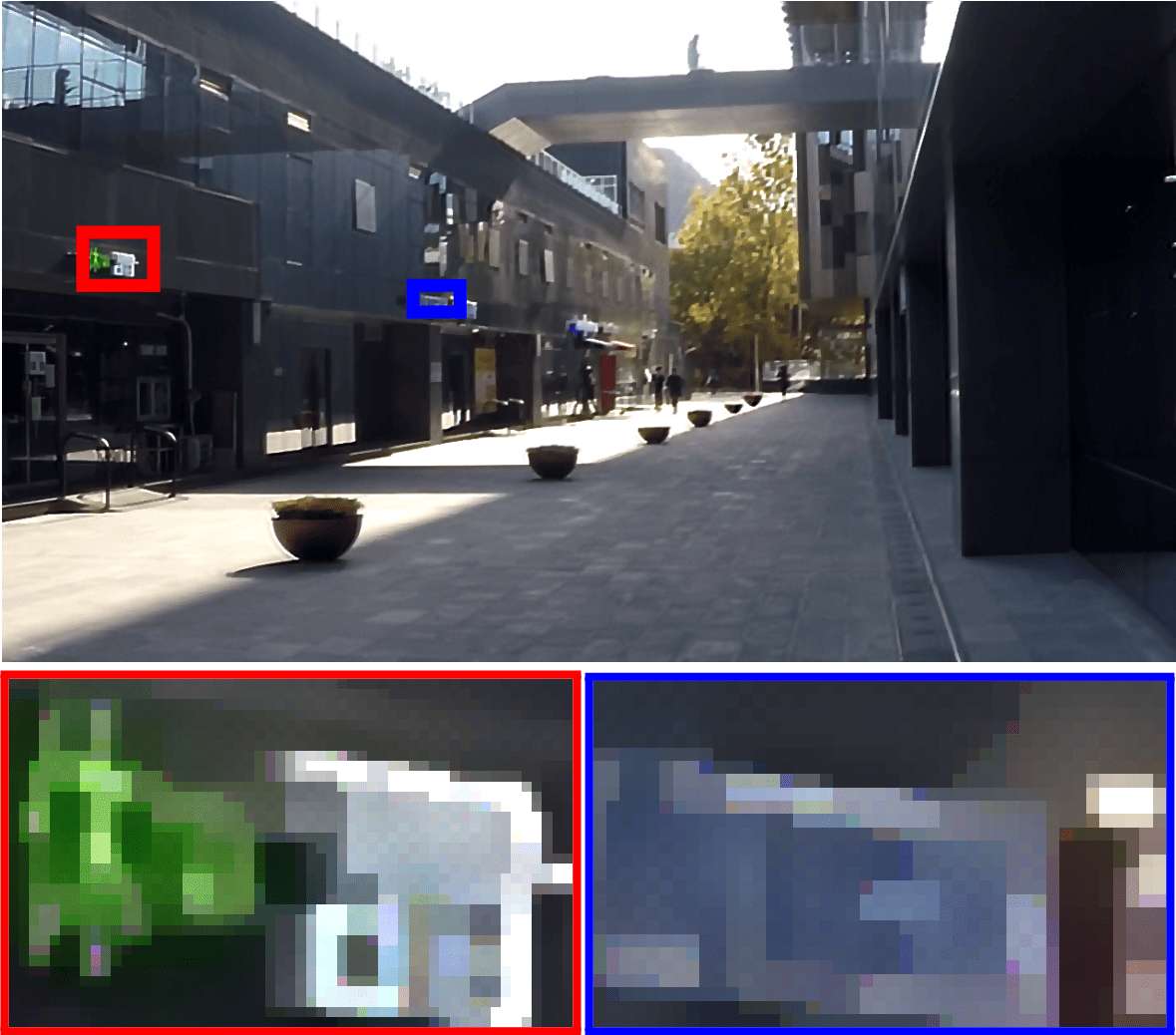}}
 \caption{ Chen \emph{et al.} \cite{chen2021blind}\\ \textbf{PSNR = 24.17, SSIM = 0.86}}
\end{subfigure}
\begin{subfigure}{1.5in}
{\includegraphics[width=1.5in,height=1.5in,keepaspectratio]{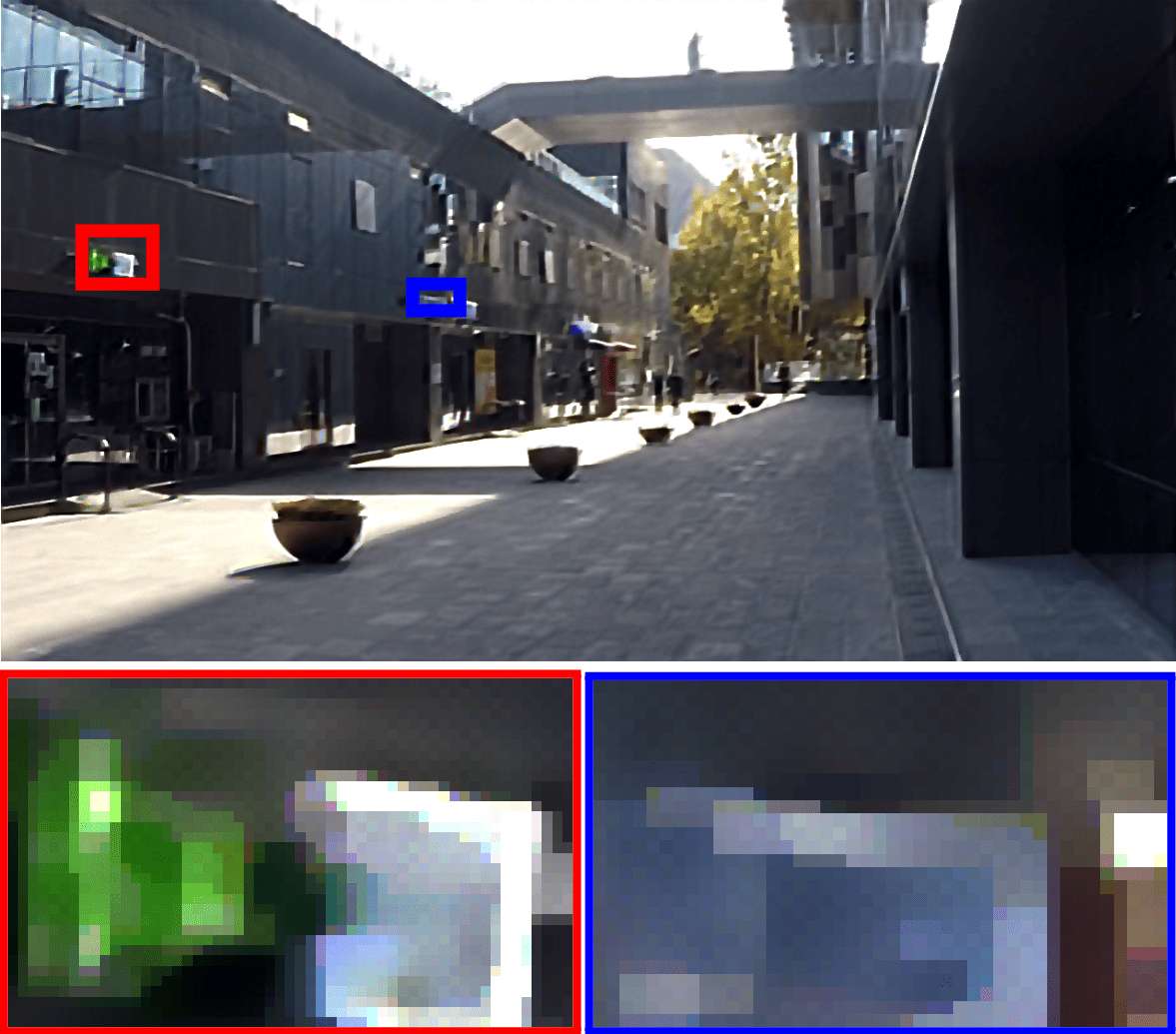}}
 \caption{ Hu \emph{et al.} \cite{hu2014deblurring}\\ \textbf{PSNR = 15.04, SSIM = 0.64}}
\end{subfigure}
\caption{Qualitative and quantitative comparison of the proposed method with state-of-the-art saturation aware deblurring techniques. The PSNR and SSIM values are presented for the entire dataset, while the displayed image is a representative sample for visual demonstration of the results on real data.}
\label{real_dataset}
\end{figure*} 
comparing it against a broad range of both saturation-specific and general-purpose image deblurring techniques to ensure coverage of the latest advancements in image restoration.\\ Notably, our comparisons include state-of-the-art deep learning \cite{9478009} based methods. To further assess the robustness of the proposed approach, we varied the degree of blur in the test dataset \cite{hanji2021hdr4cv} by merging two, three, four, and five consecutive frames, respectively, as shown in Figure \ref{blur_view}. \\
The results in Figure \ref{varying_blur} clearly demonstrate that the proposed method generally outperforms the diverse range of existing deblurring techniques. It is also worth noting that while the Structural Similarity Index (SSIM) significantly favors the proposed method compared to the second-best-performing technique, the gap in performance is less pronounced when evaluated using the Peak Signal-to-Noise Ratio (PSNR).  
SSIM measures perceived structural changes in an image, while PSNR quantifies pixel intensity differences. By weighting PSNR with SSIM, the SSIM-Weighted PSNR metric emphasizes scenarios where perceptual quality is as critical as signal fidelity 
\begin{equation}
   \text{SSIM-Weighted PSNR} = \text{PSNR} \times \text{SSIM}
\end{equation}
Similarly, The geometric mean is sensitive to low values. If either SSIM or PSNR is significantly degraded, the geometric mean will also reflect this degradation. Additionally, geometric mean inherently handles scale differences better than arithmetic means or direct comparison, as it multiplies the values and normalizes them through the root.
\begin{equation}
\text{Geometric Mean} =  \sqrt{\text{SSIM} \cdot \text{PSNR}}
\end{equation}
Together, these metrics provide a comprehensive view of image restoration performance, capturing both signal fidelity and perceptual quality, as shown in Table \ref{varying_blur_metrics}.
\begin{figure*}[!h]
\centering
\begin{subfigure}{1.36in}
{\includegraphics[width=1.36in,height=1.36in,keepaspectratio]{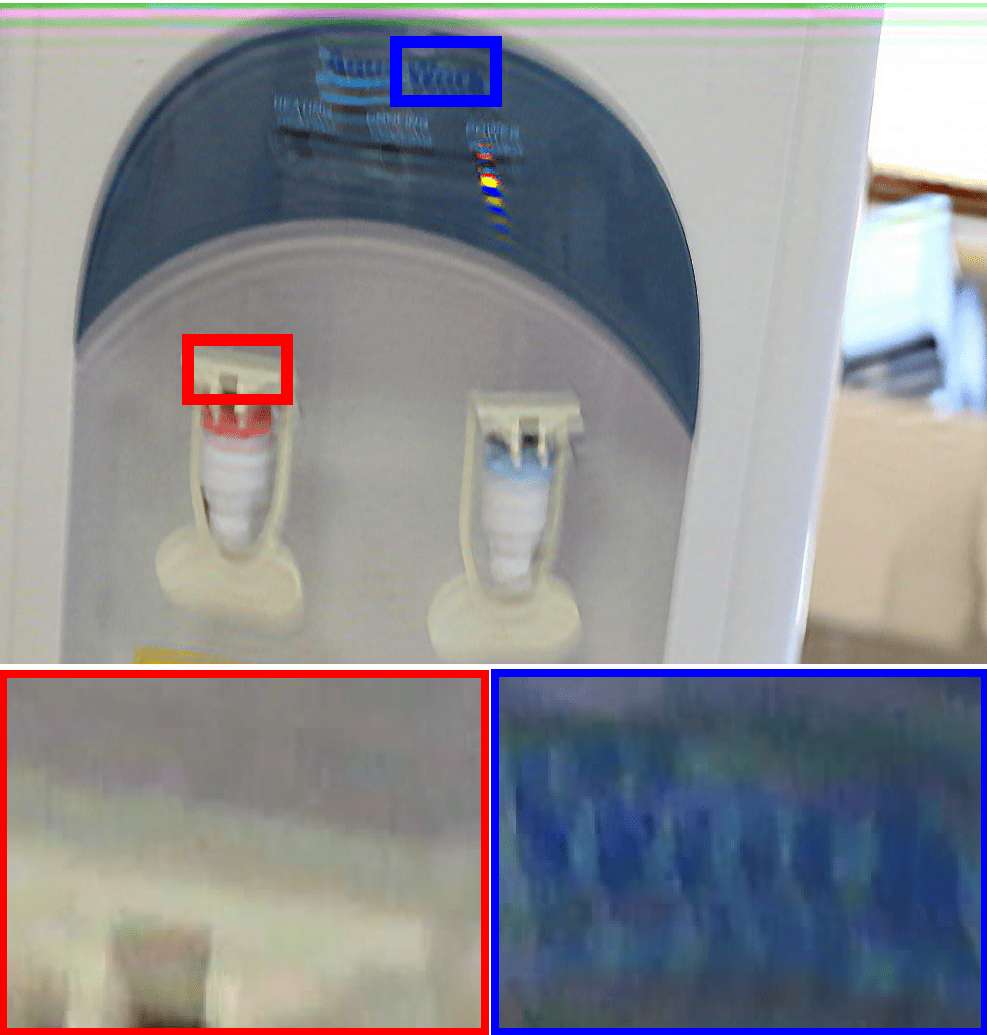}}
 \caption{ \cite{xu2010two} W/ Proposed }
\end{subfigure}
\begin{subfigure}{1.35in}
{\includegraphics[width=1.35in,height=1.35in,keepaspectratio]{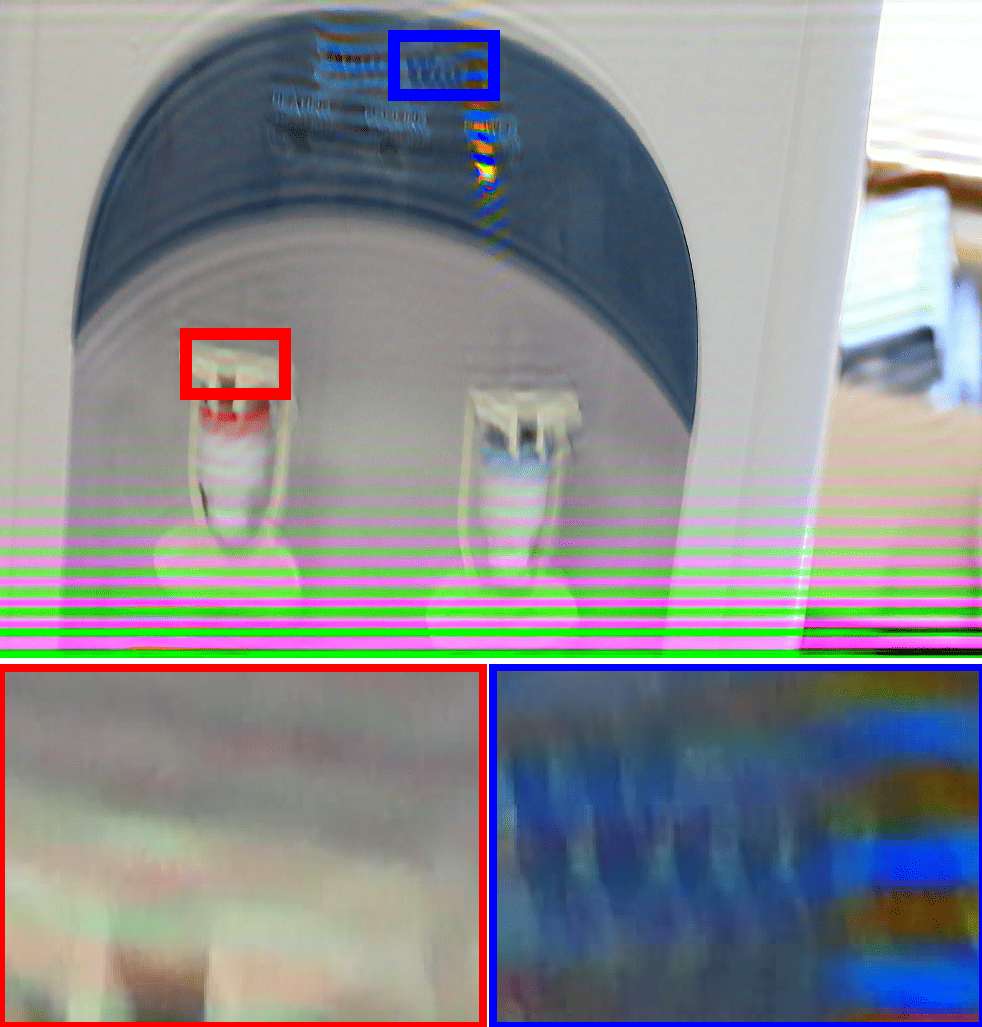}}
 \caption{ \cite{xu2010two}}
\end{subfigure}
\begin{subfigure}{1.35in}
{\includegraphics[width=1.35in,height=1.35in,keepaspectratio]{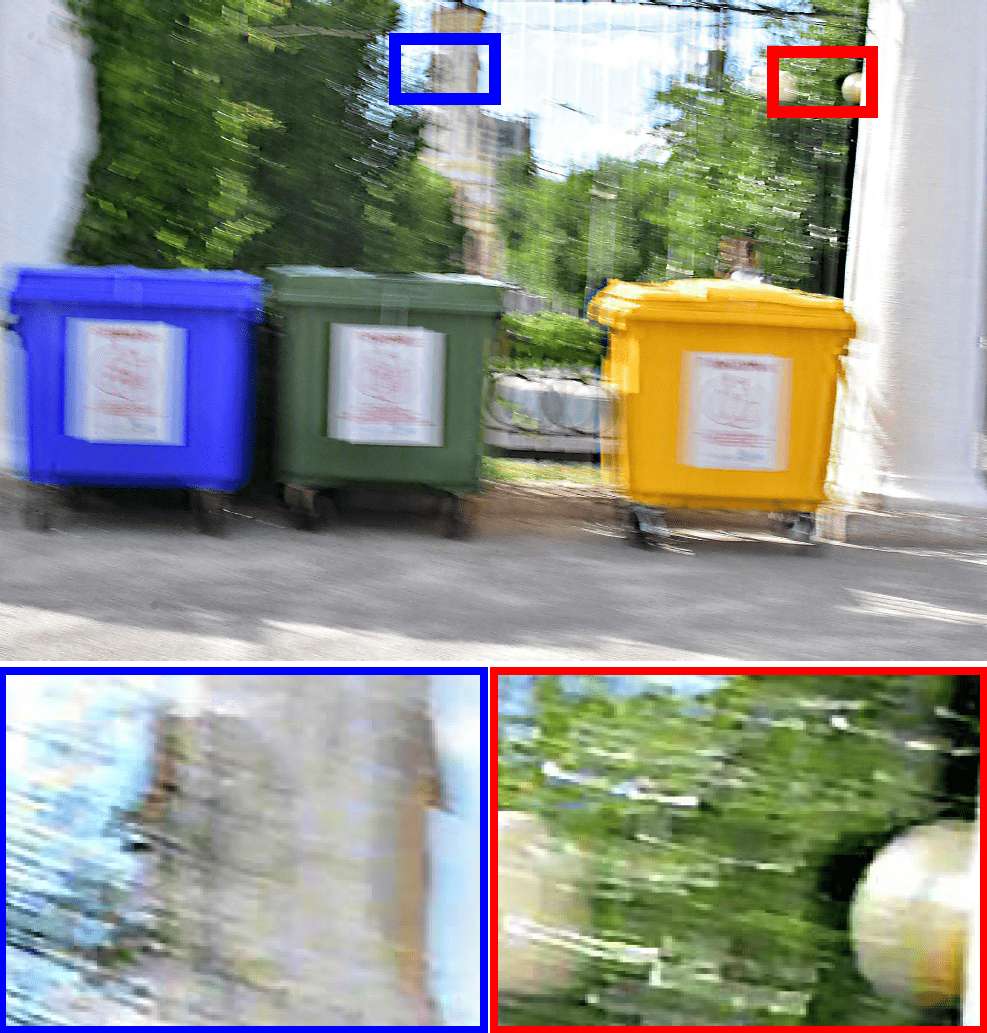}}
 \caption{\cite{xu2010two} W/ Proposed}
\end{subfigure}
\begin{subfigure}{1.35in}
{\includegraphics[width=1.35in,height=1.35in,keepaspectratio]{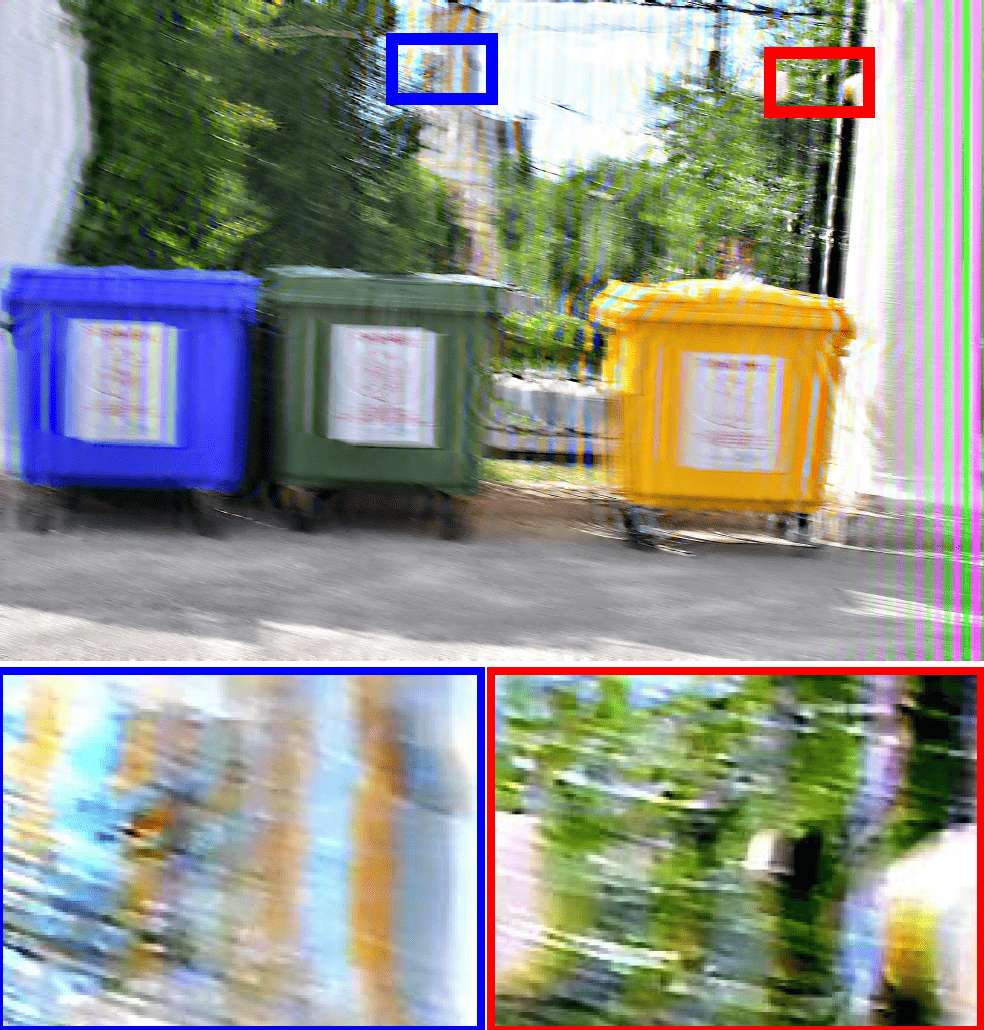}}
 \caption{\cite{xu2010two}}
\end{subfigure}
\caption{Quantitative and Qualitative comparison of deblurring results using the Xu et al. \cite{xu2010two} method with and without integration into the proposed saturation-aware framework on \cite{blur_dataset} dataset. (\cite{xu2010two} W/ Proposed) PSNR = 20.07. (\cite{xu2010two}) PSNR= 17.23.)  PSNR is calculated after the geometric registration of the deblurred image with ground truth (GT), the original dataset has visible misalignment of the degraded and GT images. }
\label{sat_effect}
\end{figure*}
The proposed method demonstrates holistic superiority compared to both saturation-specific and general-purpose learning and non-learning-based deblurring techniques.\\
We further demonstrate the effectiveness of the proposed method using a dataset captured in a real-world environment \cite{S.Nah}. Specifically, we selected the test sequence $"GOPR0881\_11\_01"$  which contains significant saturated regions in every image of the sequence and offers a high dynamic range. In this experiment, the proposed framework was used alongside an additional image deblurring technique \cite{wen2020simple} to illustrate that the framework's strength is not derived from any particular deblurring algorithm. Instead, it shows that any blind image deblurring algorithm can benefit from the proposed approach when handling saturated images. Figure \ref{real_dataset} demonstrates that the proposed framework, when used with two different deblurring methods, outperforms state-of-the-art saturation-aware deblurring techniques both qualitatively and quantitatively.\\
The two methods integrated with the proposed saturation-aware framework demonstrate significant effectiveness in handling saturation, resulting in effective deblurring as noted by their  PSNR and SSIM scores. Visually, it preserves fine details and structural integrity more effectively, while alternative methods exhibit notable artifacts and reduced clarity.\\
\begin{figure*}[!h]
\centering
\begin{subfigure}{1.1in}
{\includegraphics[width=1.1in,height=1.1in,keepaspectratio]{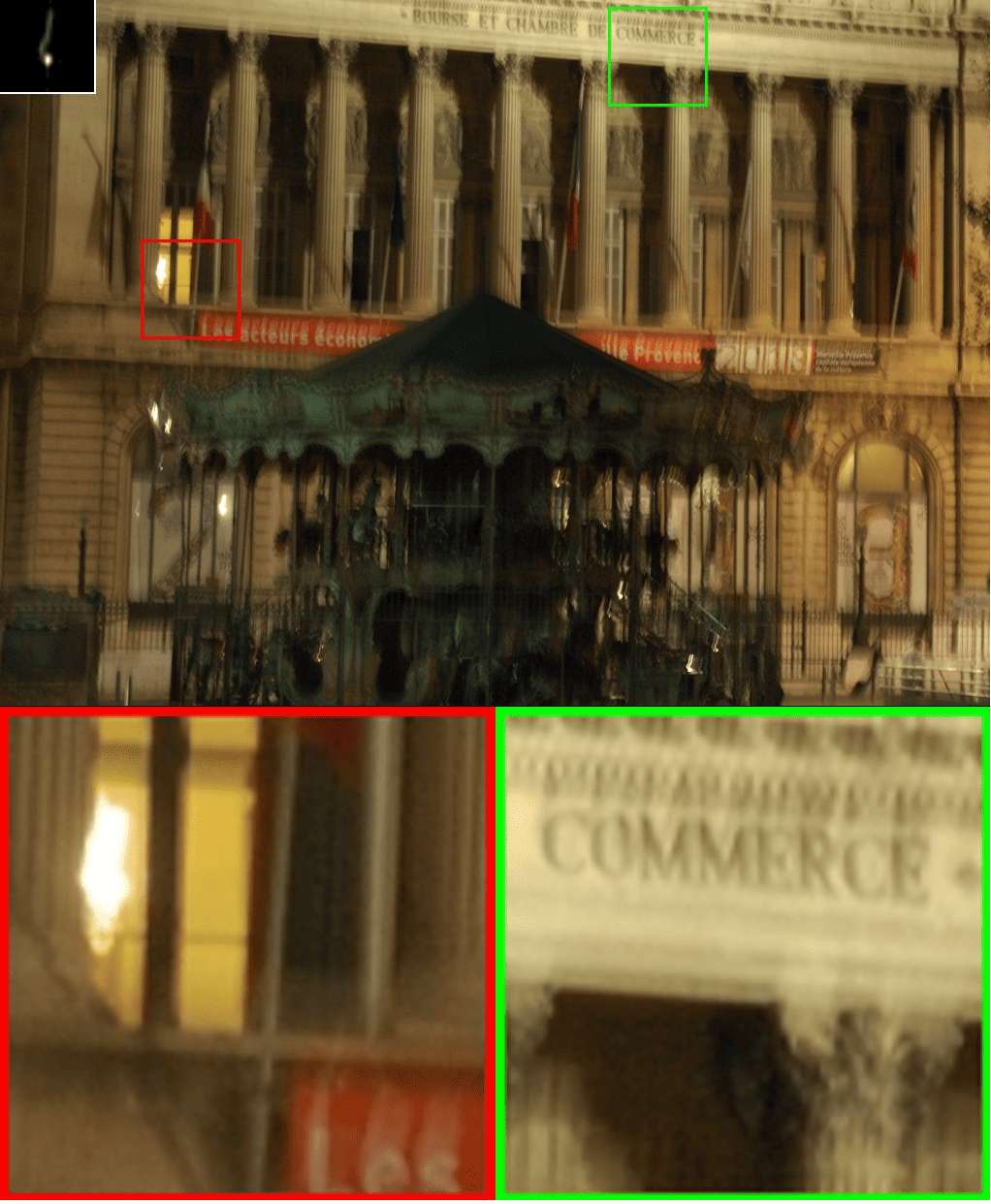}}
 \caption{Input}
\end{subfigure}\hfill
\begin{subfigure}{1.1in}
{\includegraphics[width=1.1in,height=1.1in,keepaspectratio]{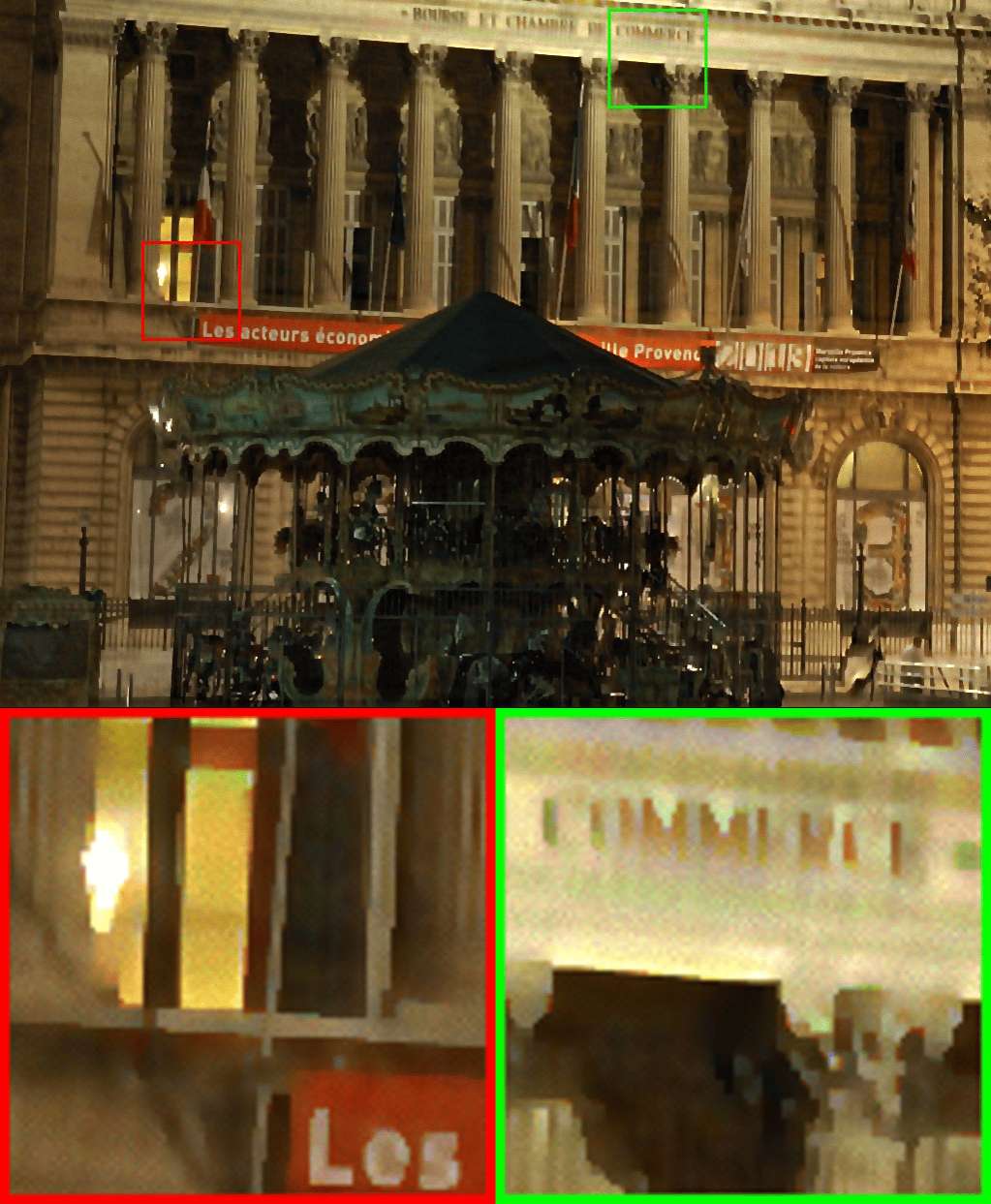}}
 \caption{chen \emph{et al.}\cite{chen2021blind}}
\end{subfigure}\hfill
\begin{subfigure}{1.1in}
{\includegraphics[width=1.1in,height=1.1in,keepaspectratio]{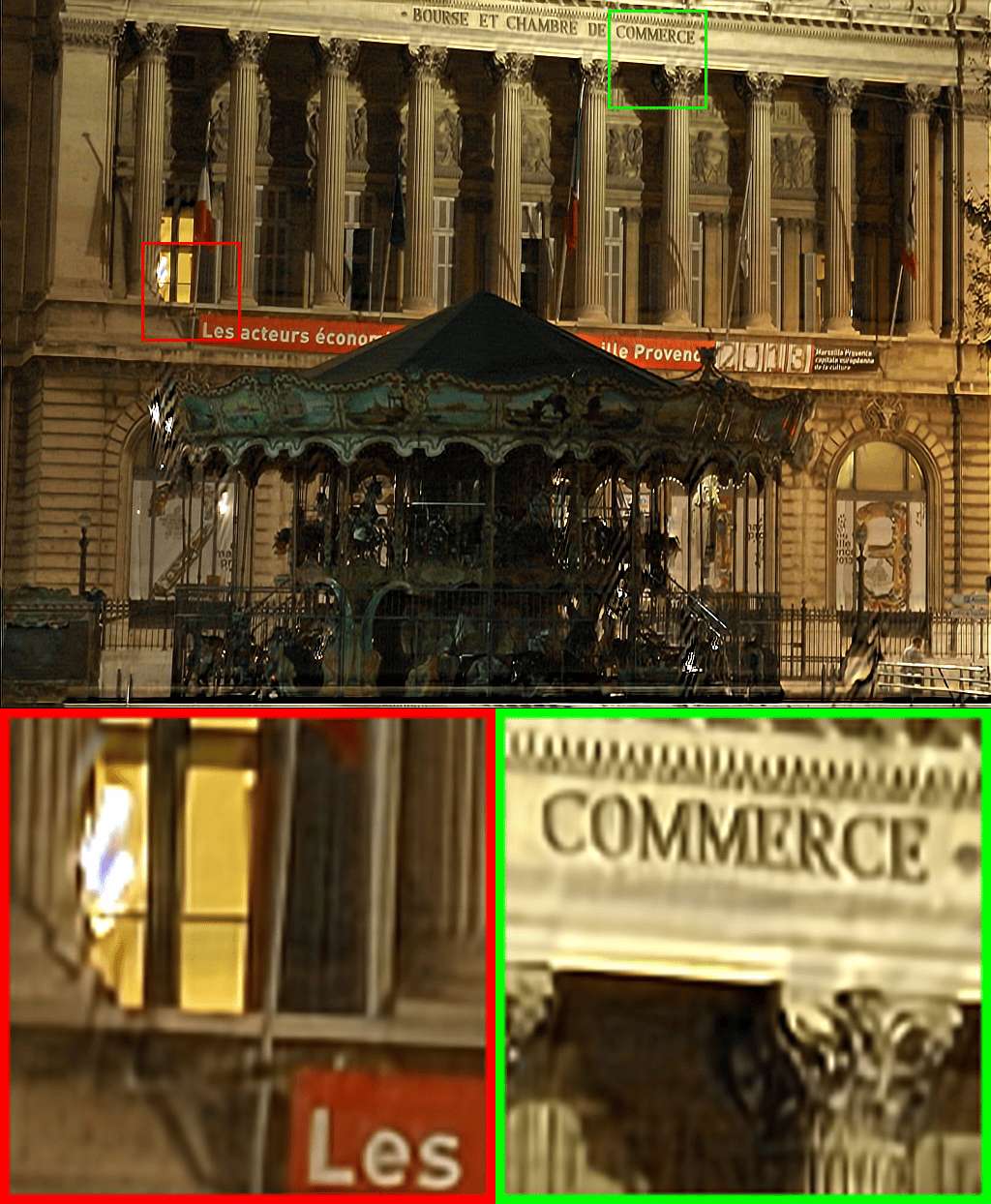}}
 \caption{Zhang \emph{et al.} \cite{zhang2017learning}}
\end{subfigure}\hfill
\begin{subfigure}{1.1in}
{\includegraphics[width=1.1in,height=1.1in,keepaspectratio]{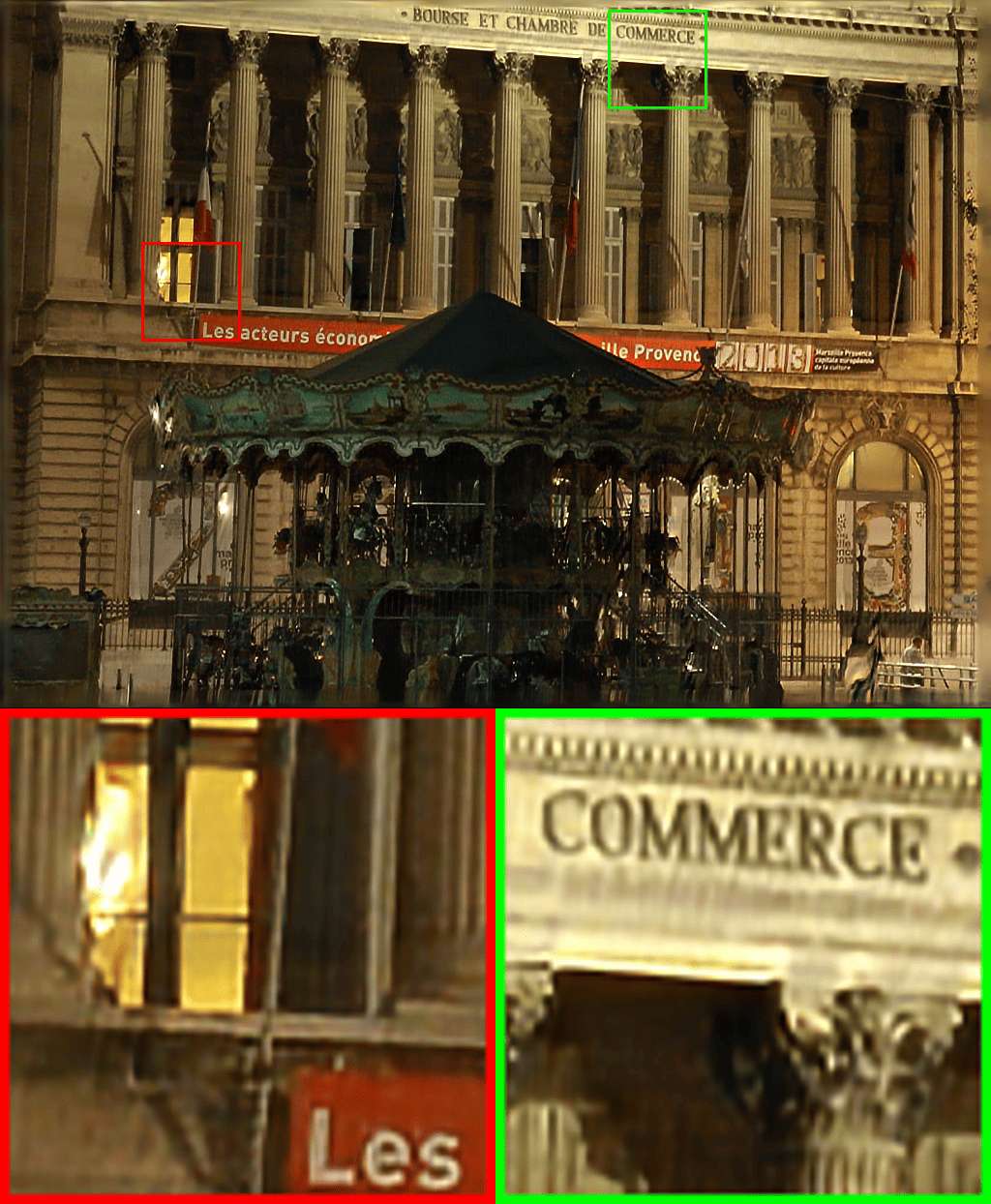}}
 \caption{chen \emph{et al.} \cite{chen2021learning}}
\end{subfigure}\hfill
\begin{subfigure}{1.1in}
{\includegraphics[width=1.1in,height=1.1in,keepaspectratio]{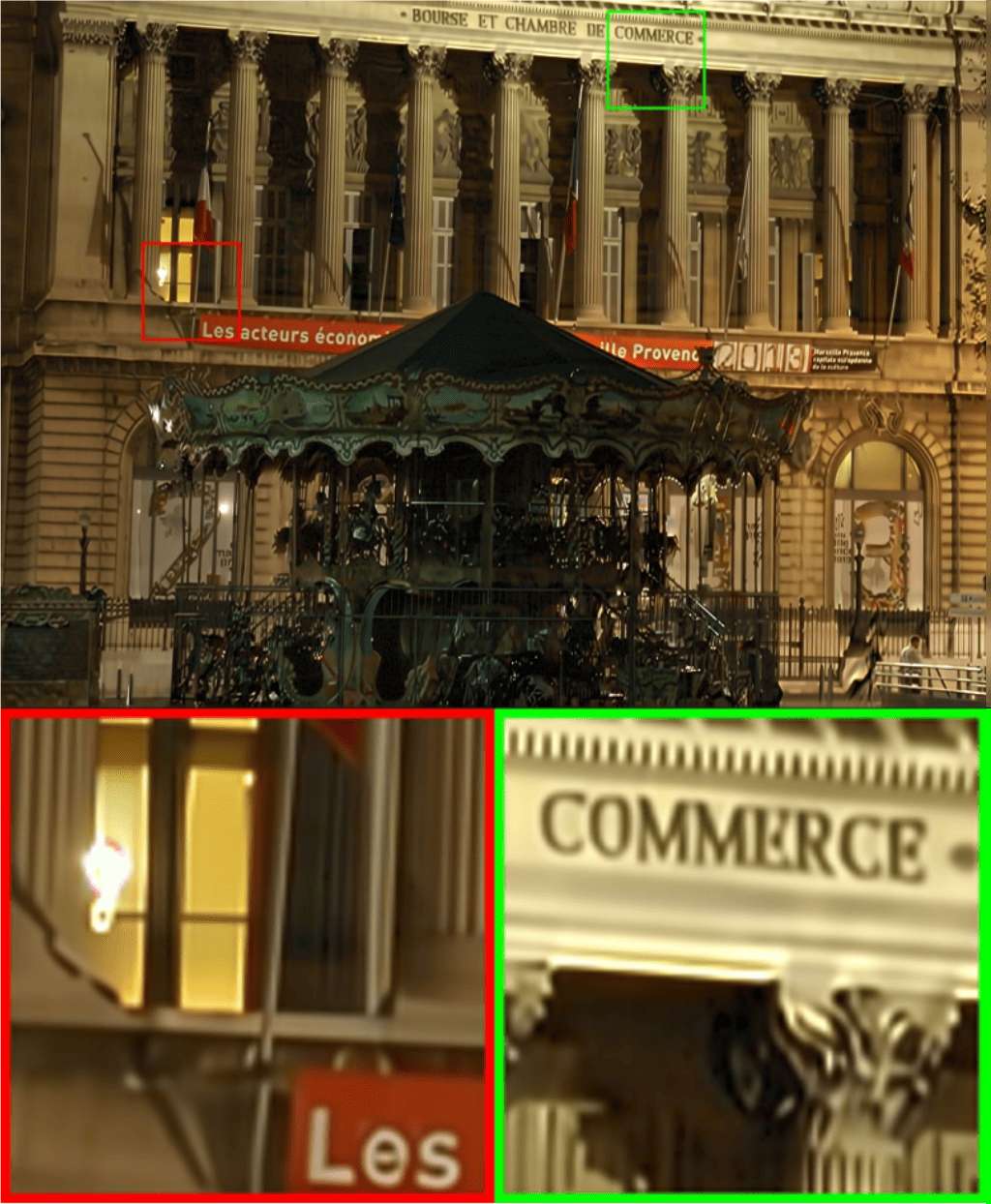}}
 \caption{Shu \emph{et al.} \cite{shu2024deep}}
\end{subfigure}\hfill
\begin{subfigure}{1.1in}
{\includegraphics[width=1.1in,height=1.1in,keepaspectratio]{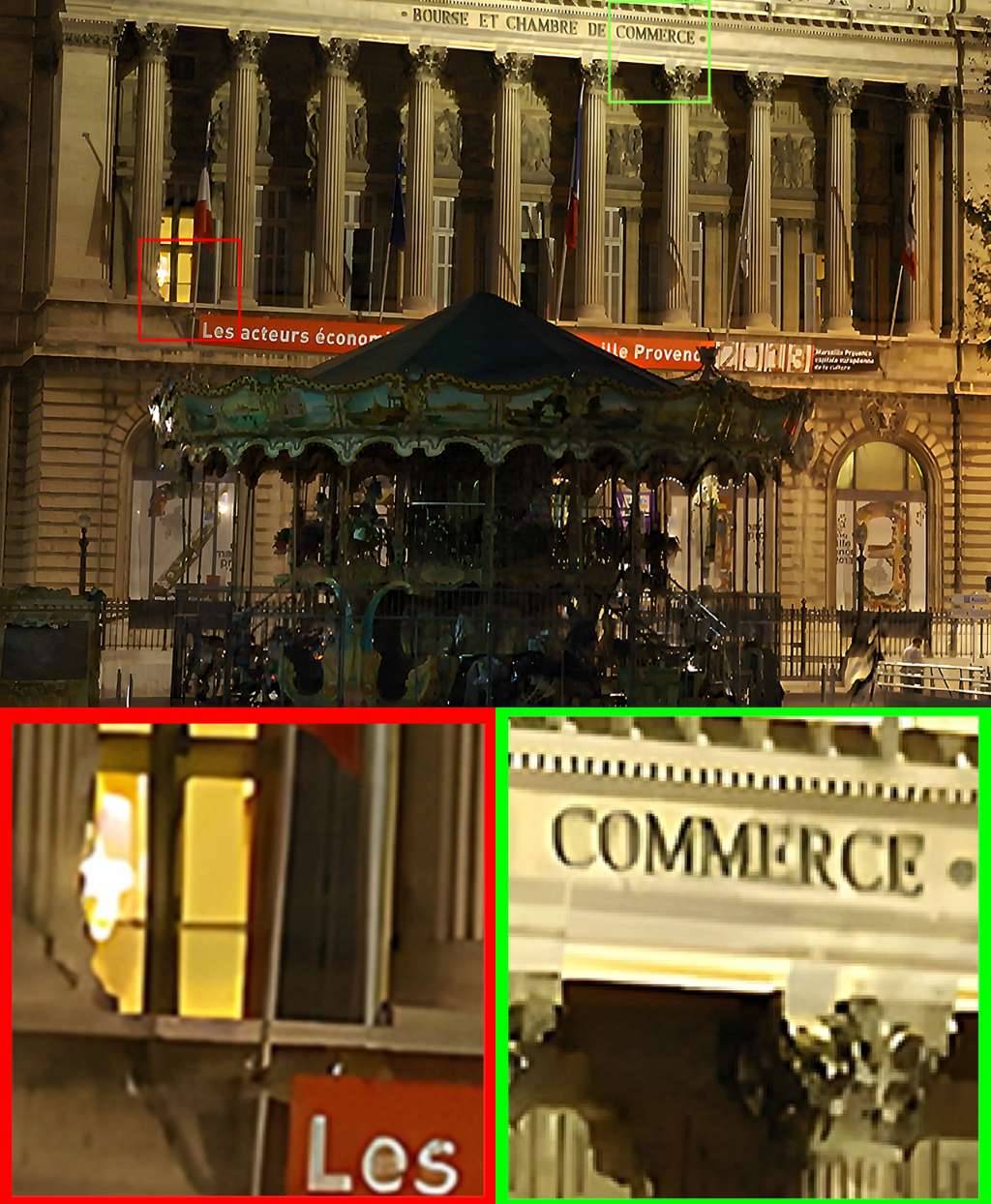}}
 \caption{Proposed}
\end{subfigure}\\
\caption{Comparative evaluation of state-of-the-art deblurring methods and the proposed approach on a real blurry image with minimal saturated regions.}
\label{mybuilding}
\end{figure*}
\begin{figure*}[!h]
\centering
\begin{subfigure}{1.1in}
{\includegraphics[width=1.1in,height=1.1in,keepaspectratio]{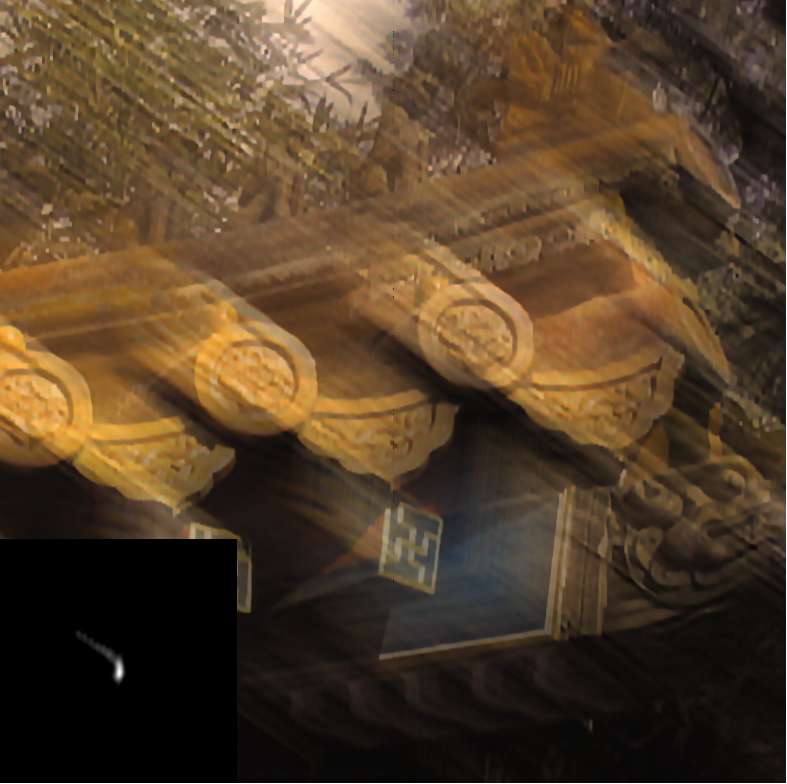}}
 \caption{Dong \emph{et al.}\cite{gong2016blind}}
\end{subfigure}\hfill
\begin{subfigure}{1.1in}
{\includegraphics[width=1.1in,height=1.1in,keepaspectratio]{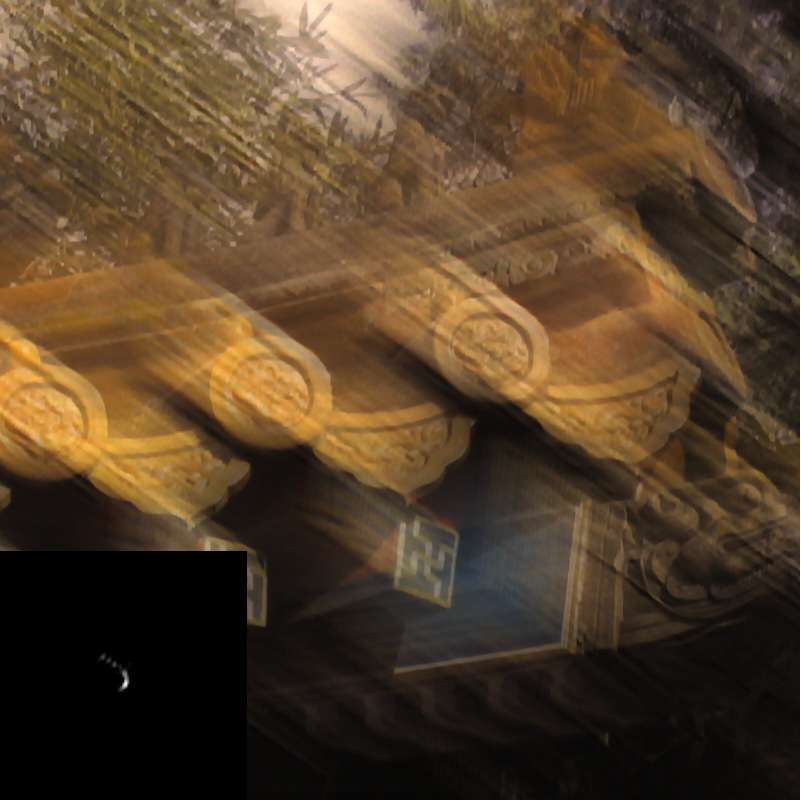}}
 \caption{chen \emph{et al.}\cite{chen2020oid}}
\end{subfigure}\hfill
\begin{subfigure}{1.1in}
{\includegraphics[width=1.1in,height=1.1in,keepaspectratio]{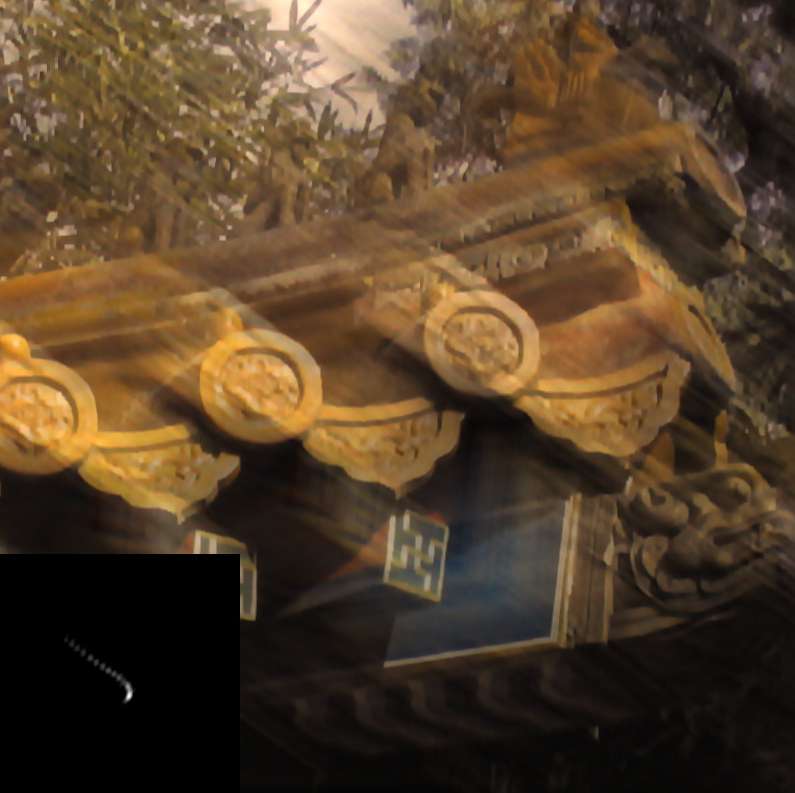}}
 \caption{Liu \emph{et al.} \cite{liu2019surface}}
\end{subfigure}\hfill
\begin{subfigure}{1.1in}
{\includegraphics[width=1.1in,height=1.1in,keepaspectratio]{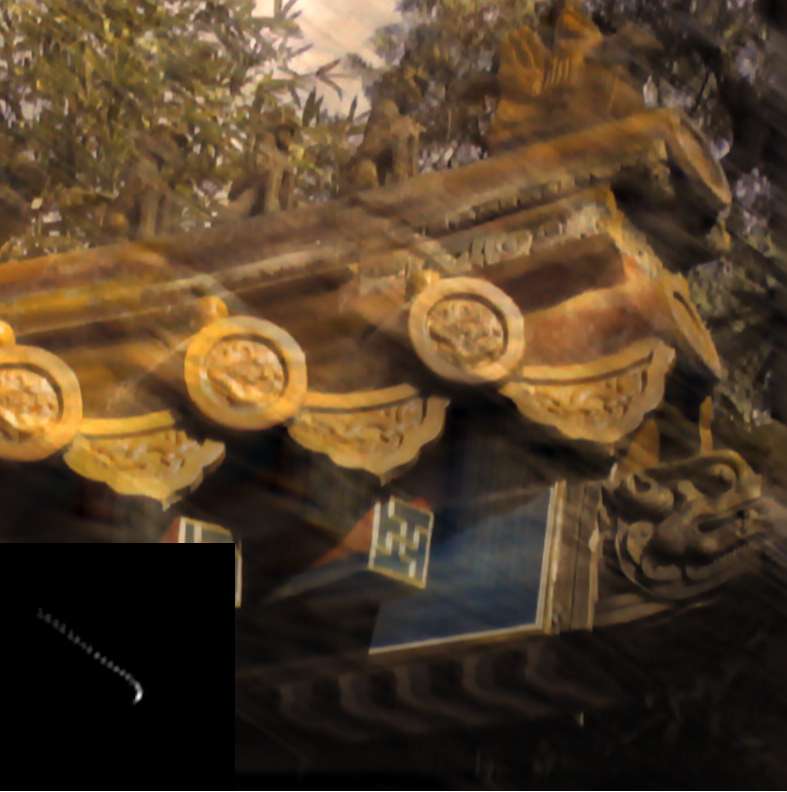}}
 \caption{Zhang \emph{et al.} \cite{zhang2022pixel}}
\end{subfigure}\hfill
\begin{subfigure}{1.1in}
{\includegraphics[width=1.1in,height=1.1in,keepaspectratio]{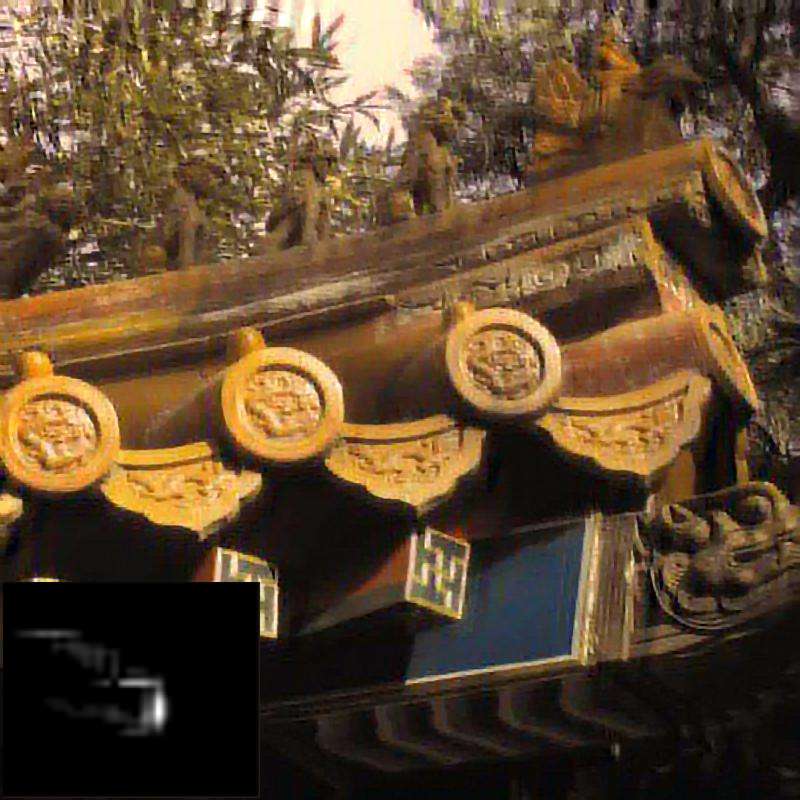}}
 \caption{Proposed}
\end{subfigure}\\
\caption{Visual comparison of state-of-the-art deblurring methods and the proposed approach on a blurry image with minimal saturated regions.}
\label{building}
\end{figure*}
Figure \ref{sat_effect} further illustrates that the proposed framework's effectiveness stems from its ability to handle image saturation, demonstrating its potential adaptability to any standard blind image deblurring method. We present two randomly chosen images from the publicly available \cite{blur_dataset} mobile phone dataset to visually compare the deblurring performance of \cite{xu2010two} when utilized within the proposed saturation-aware deblurring framework versus its direct application without the framework.\\
When \cite{xu2010two} is incorporated into the framework, the method significantly reduces ringing artifacts and better recovers fine details. In contrast, the direct application of the deblurring algorithm results in severe artifacts and loss of detail, highlighting the effectiveness of the proposed approach in enhancing image restoration.
In Fig. \ref{mybuilding} and \ref{building}, we qualitatively evaluate our proposed framework on a real blurry image from \cite{shu2024deep} and \cite{zhang2022pixel}, respectively, which contains relatively small regions affected by blur. We demonstrate that saturation, even when impacting fewer pixels, can significantly degrade the performance of image restoration techniques. For the results presented in this figure, we followed \cite{shu2024deep} in estimating the blur kernel using the method described in \cite{pan2016robust}.\color{black}
\section{Conclusion}
This work presents a targeted solution for the significant challenge of deblurring images impacted by saturation, a condition prevalent in high dynamic range and low-light settings. By introducing a saturation-aware framework that leverages the Light Spread Function (LSF) and a segmentation approach to accurately model saturated regions, the proposed method demonstrates substantial improvements in image restoration accuracy. Experimental results highlight the framework’s ability to mitigate artifacts such as ringing, outperforming existing methods in both synthetic and real-world datasets. The observed adaptability of the proposed approach underscores its potential utility across a broad spectrum of image deblurring scenarios, irrespective of the specific deblurring algorithm employed. These findings emphasize the importance of addressing saturation as a critical factor in image restoration.
\section{Limitations \& Future Work}
Our approach assumes the availability of sufficient dark pixels and minimally blurred patches near saturated regions. In extreme blur, reliably finding such patches becomes difficult; in these cases the pipeline effectively reduces to standard blind deblurring, and recovery is inherently challenging. Likewise, under extreme saturation (e.g., when a large portion of the image is clipped), the information loss at the sensor limits any method’s effectiveness, including ours.
Going forward, we will (i) investigate asymmetric / field-dependent LSF models to assess performance variations under different optical conditions; and (ii) conduct a PSF-initialization sensitivity study for the blur-proxy step (multi-start inits and robust aggregation) and report its impact on patch ranking and final quality. Although the framework currently supports both learning and non-learning deblurring back-ends, we also plan to release a Python implementation that integrates state-of-the-art deep learning methods within the same saturation-aware pipeline. \color{black}
%
%
 \bibliographystyle{IEEEbib}
\bibliography{strings.bib}

\clearpage

\end{document}